\documentclass[10pt,twocolumn,letterpaper]{article}

\usepackage[pagenumbers]{wacv} % To force page numbers, e.g. for an arXiv version

\usepackage{algorithm}
\usepackage{graphicx}
\usepackage{amsmath}
\usepackage{amssymb}
\usepackage{booktabs}
\usepackage{tabularx}
\usepackage{listings}
\usepackage[accsupp]{axessibility}

\usepackage[pagebackref,breaklinks,colorlinks]{hyperref}

\usepackage[capitalize]{cleveref}
\crefname{section}{Sec.}{Secs.}
\Crefname{section}{Section}{Sections}
\Crefname{table}{Table}{Tables}
\crefname{table}{Tab.}{Tabs.}

\def\wacvPaperID{2237} % *** Enter the WACV Paper ID here
\def\confName{WACV}
\def\confYear{2024}

\begin{document}

\title{Common Diffusion Noise Schedules and Sample Steps are Flawed}

\author{Shanchuan Lin \quad Bingchen Liu \quad Jiashi Li \quad Xiao Yang \\
ByteDance Inc.\\
{\tt\small \{peterlin,bingchenliu,lijiashi,yangxiao.0\}@bytedance.com}
}
\maketitle

%%%%%%%%% ABSTRACT
\begin{abstract}
We discover that common diffusion noise schedules do not enforce the last timestep to have zero signal-to-noise ratio (SNR), and some implementations of diffusion samplers do not start from the last timestep. Such designs are flawed and do not reflect the fact that the model is given pure Gaussian noise at inference, creating a discrepancy between training and inference. We show that the flawed design causes real problems in existing implementations. In Stable Diffusion, it severely limits the model to only generate images with medium brightness and prevents it from generating very bright and dark samples. We propose a few simple fixes: (1) rescale the noise schedule to enforce zero terminal SNR; (2) train the model with v prediction; (3) change the sampler to always start from the last timestep; (4) rescale classifier-free guidance to prevent over-exposure. These simple changes ensure the diffusion process is congruent between training and inference and allow the model to generate samples more faithful to the original data distribution.
\end{abstract}

\let\thefootnote\relax\footnotetext{Model: \href{https://huggingface.co/ByteDance/sd2.1-base-zsnr-laionaes5}{https://huggingface.co/ByteDance/sd2.1-base-zsnr-laionaes5}}

\vspace{-6pt}

%%%%%%%%% BODY TEXT
\section{Introduction}
\label{sec:intro}

Diffusion models \cite{sohldickstein2015deep,ho2020denoising} are an emerging class of generative models that have recently grown in popularity due to their capability to generate diverse and high-quality samples. Notably, an open-source implementation, Stable Diffusion \cite{rombach2021highresolution}, has been widely adopted and referenced. However, the model, up to version 2.1 at the time of writing, always generates images with medium brightness. The generated images always have mean brightness around 0 (with a brightness scale from -1 to 1) even when paired with prompts that should elicit much brighter or darker results. Moreover, the model fails to generate correct samples when paired with explicit yet simple prompts such as ``Solid black color'' or ``A white background'', etc.

\begin{figure}[t]
     \centering
     \begin{subfigure}[b]{0.495\linewidth}
         \includegraphics[width=\textwidth]{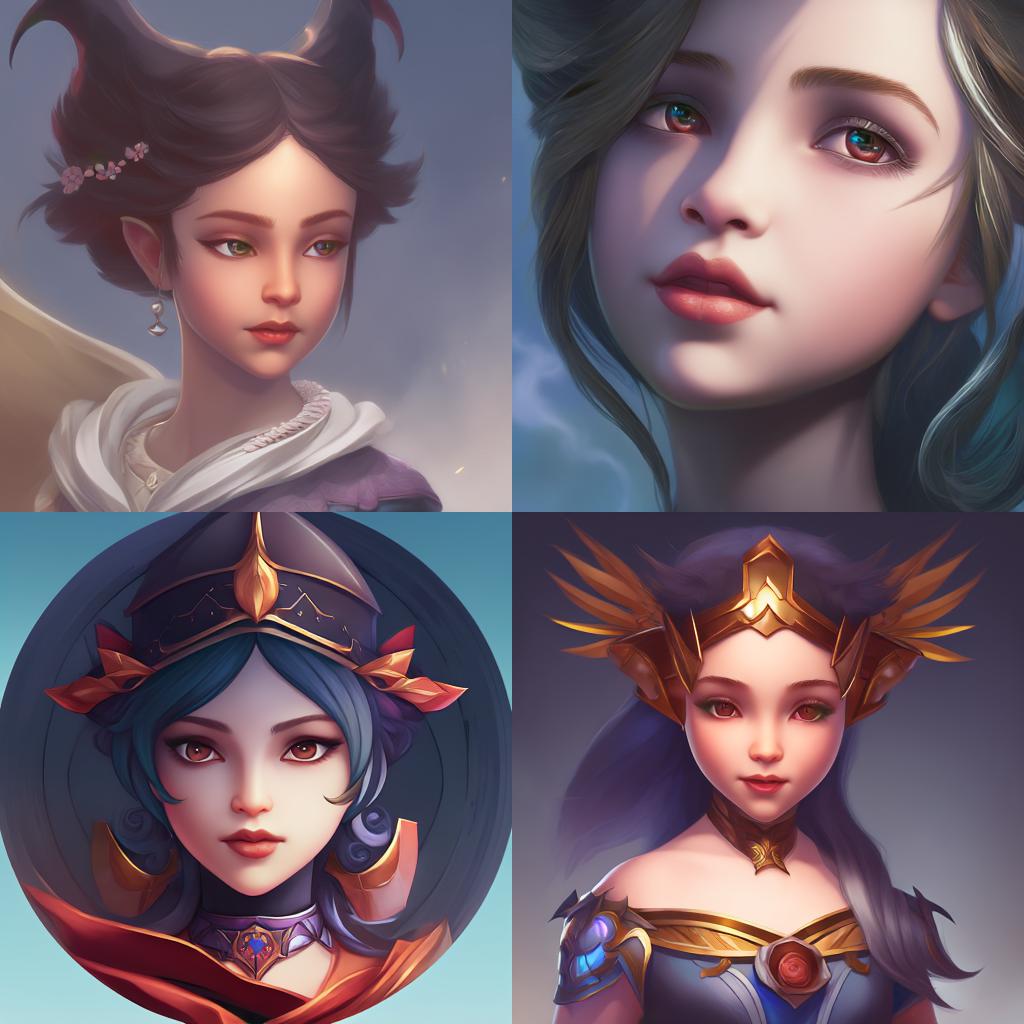}
         \caption{Flawed}
     \end{subfigure}
     \begin{subfigure}[b]{0.495\linewidth}
         \includegraphics[width=\textwidth]{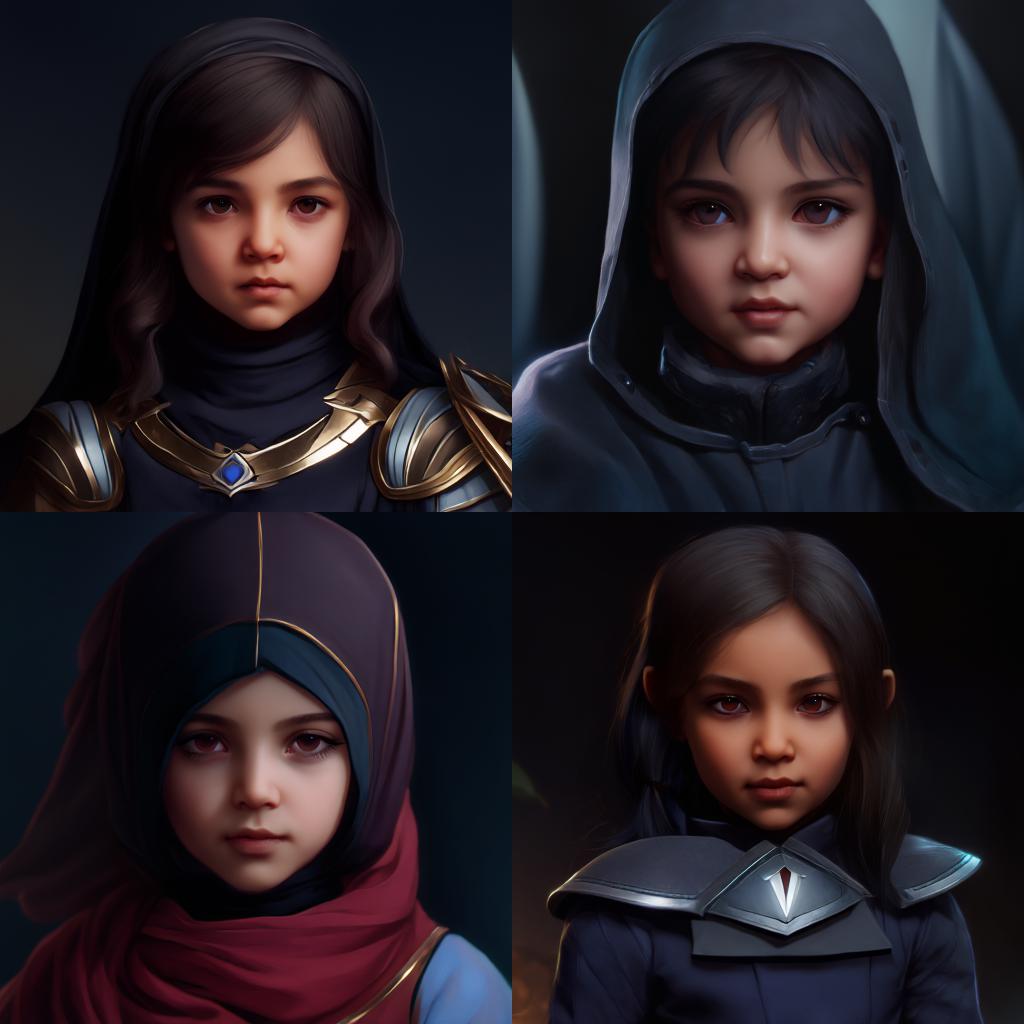}
         \caption{Corrected}
     \end{subfigure}
    \caption{Stable Diffusion uses a flawed noise schedule and sample steps which severely limit the generated images to have plain medium brightness. After correcting the flaws, the model is able to generate much darker and more cinematic images for prompt: ``Isabella, child of dark, [...] ''. Our fix allows the model to generate the full range of brightness.}
    \label{fig:intro}
    \vspace{-12pt}
\end{figure}

We discover that the root cause of the issue resides in the noise schedule and the sampling implementation. Common noise schedules, such as linear \cite{ho2020denoising} and cosine \cite{nichol2021improved} schedules, do not enforce the last timestep to have zero signal-to-noise ratio (SNR). Therefore, at training, the last timestep does not completely erase all the signal information. The leaked signal contains some of the lowest frequency information, such as the mean of each channel, which the model learns to read and respect for predicting the denoised output. However, this is incongruent with the inference behavior. At inference, the model is given pure Gaussian noise at its last timestep, of which the mean is always centered around zero. This falsely restricts the model to generating final images with medium brightness. Furthermore, newer samplers no longer require sampling of all the timesteps. However, implementations such as DDIM \cite{song2022denoising} and PNDM \cite{liu2022pseudo} do not properly start from the last timestep in the sampling process, further exacerbating the issue.

We argue that noise schedules should always ensure zero SNR on the last timestep and samplers should always start from the last timestep to correctly align diffusion training and inference. We propose a simple way to rescale existing schedules to ensure ``zero terminal SNR'' and propose a new classifier-free guidance \cite{ho2022classifierfree} rescaling technique to solve the image over-exposure problem encountered as the terminal SNR approaches zero.

We train the model on the proposed schedule and sample it with the corrected implementation. Our experimentation shows that these simple changes completely resolve the issue. These flawed designs are not exclusive to Stable Diffusion but general to all diffusion models. We encourage future designs of diffusion models to take this into account.

%-------------------------------------------------------------------------
\section{Background}

Diffusion models \cite{sohldickstein2015deep,ho2020denoising} involve a forward and a backward process. The forward process destroys information by gradually adding Gaussian noise to the data, commonly according to a non-learned, manually-defined variance schedule $\beta_1,\dots,\beta_T$. Here we consider the discrete and variance-preserving formulation, defined as:

\begin{equation}
q(x_{1:T}|x_0) := \prod^{T}_{t = 1} q(x_t|x_{t-1})    
\end{equation}

\begin{equation}
q(x_t|x_{t-1}) := \mathcal{N}(x_t; \sqrt{1-\beta_t} x_{t-1}, \beta_t \mathbf{I})    
\end{equation}

The forward process allows sampling $x_t$ at an arbitrary timestep $t$ in closed form. Let $\alpha_t := 1 - \beta_t$ and $\bar{\alpha}_t := \prod^{t}_{s=1} \alpha_s$:

\begin{equation}
q(x_t|x_0) := \mathcal{N}(x_t; \sqrt{\bar{a}_t}x_0, (1 - \bar{\alpha}_t)\mathbf{I})
\end{equation}

\noindent
Equivalently:

\begin{equation}
x_t := \sqrt{\bar{a}_t} x_0 + \sqrt{1 - \bar{a}_t} \epsilon, \quad \text{where } \epsilon \sim \mathcal{N}(\mathbf{0},\mathbf{I})
\label{eq:p_sample}
\end{equation}

Signal-to-noise ratio (SNR) can be calculated as:

\begin{equation}
\text{SNR}(t) := \frac{\bar{\alpha}_t}{1 - \bar{\alpha}_t}
\end{equation}

Diffusion models learn the backward process to restore information step by step. When $\beta_t$ is small, the reverse step is also found to be Gaussian:

\begin{equation}
p_{\theta}(x_{0:T}) := p(x_T) \prod^{T}_{t=1} p_{\theta}(x_{t-1}|x_t)
\end{equation}

\begin{equation}
p_{\theta} (x_{t-1}|x_t) := \mathcal{N}(x_{t-1};\Tilde{\mu}_t, \Tilde{\beta}_t\mathbf{I})
\end{equation}

Neural models are used to predict $\Tilde{\mu}_t$. Commonly, the models are reparameterized to predict noise $\epsilon$ instead, since:

\begin{equation}
\Tilde{\mu}_t := \frac{1}{\sqrt{\alpha_t}} (x_t - \frac{\beta_t}{\sqrt{1 - \bar{\alpha}_t}} \epsilon)
\end{equation}

Variance $\Tilde{\beta}_t$ can be calculated from the forward process posteriors:

\begin{equation}
\Tilde{\beta}_t := \frac{1 - \bar{\alpha}_{t-1}}{1 - \bar{\alpha}_t} \beta_t
\end{equation}

\section{Methods}

\subsection{Enforce Zero Terminal SNR}
\label{sec:enforce_zero_terminal_snr}

\begin{table*}[h!]
    \centering
    \begin{tabularx}{\textwidth}{l|X|rrr}
        \toprule
        Schedule & Definition ($i=\frac{t-1}{T-1}$) & $\text{SNR}(T)$ & $\sqrt{\bar{\alpha_T}}$ \\
        \midrule
        Linear \cite{ho2020denoising} & $\beta_t = 0.0001 \cdot (1 - i) + 0.02 \cdot i$ & 4.035993e-05 & 0.006353 \\
        Cosine \cite{nichol2021improved} & $\beta_t = \text{min}(1 - \frac{\bar{\alpha}_t}{\bar{\alpha}_{t-1}}, 0.999), \bar{\alpha}_t = \frac{f(t)}{f(0)}, f(t)=\text{cos}(\frac{i+0.008}{1+0.008} \cdot \frac{\pi}{2})^2$ & 2.428735e-09 & 4.928220e-05 \\
        Stable Diffusion \cite{rombach2021highresolution} & $\beta_t = (\sqrt{0.00085} \cdot (1 - i) + \sqrt{0.012} \cdot i)^2$ & 0.004682 & 0.068265 \\
        \bottomrule
    \end{tabularx}
    \caption{Common schedule definitions and their SNR and $\sqrt{\bar{\alpha}}$ on the last timestep. All schedules use total timestep $T=1000$. None of the schedules has zero SNR on the last timestep $t=T$, causing inconsistency in train/inference behavior.}
    \label{tab:schedule_definition}
\end{table*}

\Cref{tab:schedule_definition} shows common schedule definitions and their $\text{SNR}(T)$ and $\sqrt{\bar{\alpha}_T}$ at the terminal timestep $T=1000$. None of the schedules have zero terminal SNR. Moreover, cosine schedule deliberately clips $\beta_t$ to be no greater than 0.999 to prevent terminal SNR from reaching zero.

We notice that the noise schedule used by Stable Diffusion is particularly flawed. The terminal SNR is far from reaching zero. Substituting the value into \Cref{eq:p_sample} also reveals that the signal is far from being completely destroyed at the final timestep:

\begin{equation}
x_T = 0.068265 \cdot x_0 + 0.997667 \cdot \epsilon
\end{equation}

This effectively creates a gap between training and inference. When $t=T$ at training, the input to the model is not completely pure noise. A small amount of signal is still included. The leaked signal contains the lowest frequency information, such as the overall mean of each channel. The model subsequently learns to denoise respecting the mean from the leaked signal. At inference, pure Gaussian noise is given for sampling instead. The Gaussian noise always has a zero mean, so the model continues to generate samples according to the mean given at $t=T$, resulting in images with medium brightness. In contrast, a noise schedule with zero terminal SNR uses pure noise as input at $t=T$ during training, thus consistent with the inference behavior.

The same problem extrapolates to all diffusion noise schedules in general, although other schedules' terminal SNRs are closer to zero so it is harder to notice in practice. We argue that diffusion noise schedules must enforce zero terminal SNR to completely remove the discrepancy between training and inference. This also means that we must use variance-preserving formulation since variance-exploding formulation \cite{song2021scorebased} cannot truly reach zero terminal SNR.

We propose a simple fix by rescaling existing noise schedules under the variance-preserving formulation to enforce zero terminal SNR. Recall in \Cref{eq:p_sample} that $\sqrt{\bar{\alpha}_t}$ controls the amount of signal to be mixed in. The idea is to keep $\sqrt{\bar{\alpha}_1}$ unchanged, change $\sqrt{\bar{\alpha}_T}$ to zero, and linearly rescale $\sqrt{\bar{\alpha}_t}$ for intermediate $t \in [2,\dots,T-1]$ respectively. We find scaling the schedule in $\sqrt{\bar{\alpha}_t}$ space can better preserve the curve than scaling in $\text{SNR}(t)$ space. The PyTorch implementation is given in \Cref{algo:enforce_zero_terminal_snr}.

\begin{figure}
     \centering
     \begin{subfigure}[b]{0.235\textwidth}
         \centering
         \includegraphics[width=\textwidth]{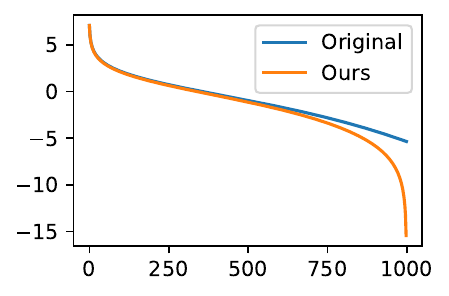}
         \caption{logSNR$(t)$}
     \end{subfigure}
     \begin{subfigure}[b]{0.235\textwidth}
         \centering
         \includegraphics[width=\textwidth]{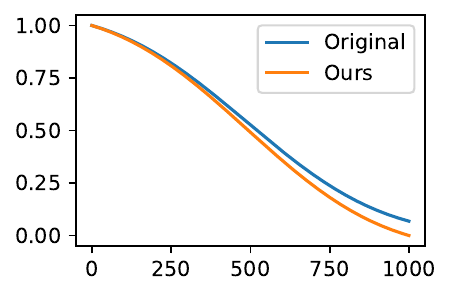}
         \caption{$\sqrt{\bar{\alpha}_t}$}
     \end{subfigure}
    \caption{Comparison between original Stable Diffusion noise schedule and our rescaled noise schedule. Our rescaled noise schedule ensures zero terminal SNR.}
    \label{fig:schedule}
\end{figure}

\definecolor{codegreen}{rgb}{0,0.6,0}
\definecolor{codegray}{rgb}{0.5,0.5,0.5}
\definecolor{codepurple}{rgb}{0.58,0,0.82}

\lstdefinestyle{mystyle}{
    commentstyle=\color{codegreen},
    keywordstyle=\color{magenta},
    numberstyle=\tiny\color{codegray},
    stringstyle=\color{codepurple},
    basicstyle=\ttfamily\footnotesize,
    breakatwhitespace=false,         
    breaklines=true,                 
    captionpos=b,                    
    keepspaces=true,                 
    numbers=left,                    
    numbersep=5pt,                  
    showspaces=false,                
    showstringspaces=false,
    showtabs=false,                  
    tabsize=2
}

\lstset{style=mystyle}

\begin{algorithm}[h!]
\begin{lstlisting}[language=Python]
def enforce_zero_terminal_snr(betas):
  # Convert betas to alphas_bar_sqrt
  alphas = 1 - betas
  alphas_bar = alphas.cumprod(0)
  alphas_bar_sqrt = alphas_bar.sqrt()

  # Store old values.
  alphas_bar_sqrt_0 = alphas_bar_sqrt[0].clone()
  alphas_bar_sqrt_T = alphas_bar_sqrt[-1].clone()
  # Shift so last timestep is zero.
  alphas_bar_sqrt -= alphas_bar_sqrt_T
  # Scale so first timestep is back to old value.
  alphas_bar_sqrt *= alphas_bar_sqrt_0 / (alphas_bar_sqrt_0 - alphas_bar_sqrt_T)

  # Convert alphas_bar_sqrt to betas
  alphas_bar = alphas_bar_sqrt ** 2
  alphas = alphas_bar[1:] / alphas_bar[:-1]
  alphas = torch.cat([alphas_bar[0:1], alphas])
  betas = 1 - alphas
  return betas
\end{lstlisting}
\caption{Rescale Schedule to Zero Terminal SNR}
\label{algo:enforce_zero_terminal_snr}
\end{algorithm}

Note that the proposed rescale method is only needed for fixing existing non-cosine schedules. Cosine schedule can simply remove the $\beta_t$ clipping to achieve zero terminal SNR. Schedules designed in the future should ensure $\beta_T = 1$ to achieve zero terminal SNR.

\subsection{Train with V Prediction and V Loss}

When SNR is zero, $\epsilon$ prediction becomes a trivial task and $\epsilon$ loss cannot guide the model to learn anything meaningful about the data.

We switch to v prediction and v loss as proposed in \cite{salimans2022progressive}:

\begin{equation}
v_t = \sqrt{\bar{\alpha}_t} \epsilon - \sqrt{1 - \bar{\alpha}_t} x_0
\end{equation}

\begin{equation}
\mathcal{L} = \lambda_t || v_t - \Tilde{v}_t ||^2_2
\end{equation}

After rescaling the schedule to have zero terminal SNR, at $t=T$, $\bar{\alpha}_T = 0$, so $v_T = x_0$. Therefore, the model is given pure noise $\epsilon$ as input to predict $x_0$ as output. At this particular timestep, the model is not performing the denoising task since the input does not contain any signal. Rather it is repurposed to predict the mean of the data distribution conditioned on the prompt.

We finetune Stable Diffusion model using v loss with $\lambda_t = 1$ and find the visual quality similar to using $\epsilon$ loss. We recommend always using v prediction for the model and adjusting $\lambda_t$ to achieve different loss weighting if desired.

\subsection{Sample from the Last Timestep}

\begin{table*}[h!]
    \centering
    \setlength{\tabcolsep}{0.31em}
    \begin{tabularx}{\textwidth}{l|l|l|rrrrrrrrrr}
        \toprule
        Type & Method & Discretization & \multicolumn{10}{l}{Timesteps (\eg $T=1000, S=10$)} \\
        \midrule
        Leading & DDIM\cite{ho2020denoising}, PNDM\cite{liu2022pseudo} & $\text{arange}(1, T+1, \text{floor}(T/S))$ & 1 & 101 & 201 & 301 & 401 & 501 & 601 & 701 & 801 & 901 \\
        Linspace & iDDPM\cite{nichol2021improved} & $\text{round}(\text{linspace}(1, T, S))$ & 1 &  112 & 223 & 334 & 445 & 556 & 667 & 778 & 889 & 1000 \\
        Trailing & DPM\cite{lu2022dpmsolver} & $\text{round}(\text{flip}(\text{arange}(T, 0, -T/S)))$ & 100 & 200 & 300 & 400 & 500 & 600 & 700 & 800 & 900 & 1000 \\
        \bottomrule
    \end{tabularx}
    \caption{Comparison between sample steps selections. $T$ is the total discrete timesteps the model is trained on. $S$ is the number of sample steps used by the sampler. We argue that sample steps should always include the last timestep $t=T$ in the sampling process. Example here uses $T=1000, S=10$ only for illustration proposes. Note that the timestep here uses range $[1,\dots,1000]$ to match the math notation used in the paper but in practice most implementations use timestep range $[0,\dots,999]$ so it should be shifted accordingly.}
    \label{tab:sampling_steps}
\end{table*}

Newer samplers are able to sample much fewer steps to generate visually appealing samples. Common practice is to still train the model on a large amount of discretized timesteps, \eg $T=1000$, and only perform a few sample steps, \eg $S=25$, at inference. This allows the dynamic change of sample steps $S$ at inference to trade-off between quality and speed.

However, many implementations, including the official DDIM \cite{song2022denoising} and PNDM \cite{liu2022pseudo} implementations, do not properly include the last timestep in the sampling process, as shown in \Cref{tab:sampling_steps}. This is also incorrect because models operating at $t<T$ are trained on non-zero SNR inputs thus inconsistent with the inference behavior. For the same reason discussed in \Cref{sec:enforce_zero_terminal_snr}, this contributes to the brightness problem in Stable Diffusion.

We argue that sampling from the last timestep in conjunction with a noise schedule that enforces zero terminal SNR is crucial. Only this way, when pure Gaussian noise is given to the model at the initial sample step, the model is actually trained to expect such input at inference.

We consider two additional ways to select sample steps in \Cref{tab:sampling_steps}. Linspace, proposed in iDDPM \cite{nichol2021improved}, includes both the first and the last timestep and then evenly selects intermediate timesteps through linear interpolation. Trailing, proposed in DPM\cite{lu2022dpmsolver}, only includes the last timestep and then selects intermediate timesteps with an even interval starting from the end. Note that the selection of the sample steps is not bind to any particular sampler and can be easily interchanged.

We find trailing has a more efficient use of the sample steps especially when $S$ is small. This is because, for most schedules, $x_1$ only differs to $x_0$ by a tiny amount of noise controlled by $\beta_1$ and the model does not perform many meaningful changes when sampled at $t = 1$, effectively making the sample step at $t = 1$ useless. We switch to trailing for future experimentation and use DDIM to match the official Stable Diffusion implementation.

Note that some sampler implementations may encounter zero division errors. The fix is provided in \Cref{sec:implementation}.

\subsection{Rescale Classifier-Free Guidance}

We find that as the terminal SNR approaches zero, classifier-free guidance \cite{ho2022classifierfree} becomes very sensitive and can cause images to be overexposed. This problem has been noticed in other works. For example, Imagen \cite{saharia2022photorealistic} uses cosine schedule, which has a near zero terminal SNR, and proposes dynamic thresholding to solve the over-exposure problem. However, the approach is designed only for image-space models. Inspired by it, we propose a new way to rescale classifier-free guidance that is applicable to both image-space and latent-space models.

\begin{equation}
x_{cfg} = x_{neg} + w (x_{pos} - x_{neg})
\label{eq:cfg}
\end{equation}

\Cref{eq:cfg} shows regular classifier-free guidance, where $w$ is the guidance weight, $x_{pos}$ and $x_{neg}$ are the model outputs using positive and negative prompts respectively. We find that when $w$ is large, the scale of the resulting $x_{cfg}$ is very big, causing the image over-exposure problem. To solve it, we propose to rescale after applying classifier-free guidance:

\begin{equation}
\sigma_{pos} = \text{std}(x_{pos}), \quad
\sigma_{cfg} = \text{std}(x_{cfg})
\label{eq:cfg_std}
\end{equation}

\begin{equation}
x_{rescaled} = x_{cfg} \cdot \frac{\sigma_{pos}}{\sigma_{cfg}}
\label{eq:cfg_rescaled}
\end{equation}

\begin{equation}
x_{final} = \phi \cdot x_{rescaled} + (1 - \phi) \cdot x_{cfg}
\label{eq:cfg_final}
\end{equation}

In \Cref{eq:cfg_std}, we compute the standard deviation of $x_{pos}, x_{cfg}$ as $\sigma_{pos}, \sigma_{cfg} \in \mathbb{R}$. In \Cref{eq:cfg_rescaled}, we rescale $x_{cfg}$ back to the original standard deviation before applying classifier-free guidance but discover that the generated images are overly plain. In \Cref{eq:cfg_final}, we introduce a hyperparameter $\phi$ to control the rescale strength. We empirically find $w=7.5, \phi=0.7$ works great. The optimized PyTorch implementation is given in \Cref{algo:cfg_rescale}.

\begin{algorithm}[h!]
\begin{lstlisting}[language=Python]
def apply_cfg(pos, neg, weight=7.5, rescale=0.7):
  # Apply regular classifier-free guidance.
  cfg = neg + weight * (pos - neg)
  # Calculate standard deviations.
  std_pos = pos.std([1,2,3], keepdim=True)
  std_cfg = cfg.std([1,2,3], keepdim=True)
  # Apply guidance rescale with fused operations.
  factor = std_pos / std_cfg
  factor = rescale * factor + (1 - rescale)
  return cfg * factor
\end{lstlisting}
\caption{Classifier-Free Guidance with Rescale}
\label{algo:cfg_rescale}
\end{algorithm}

\section{Evaluation}

We finetune Stable Diffusion 2.1-base model on Laion dataset \cite{schuhmann2022laion5b} using our fixes. Our Laion dataset is filtered similarly to the data used by the original Stable Diffusion. We use the same training configurations, i.e. batch size 2048, learning rate 1e-4, ema decay 0.9999, to train our model for 50k iterations. We also train an unchanged reference model on our filtered Laion data for a fair comparison.

\subsection{Qualitative}

\begin{figure*}[h!]
    \centering
    \captionsetup{justification=raggedright,singlelinecheck=false}
    \small
    \begin{tabularx}{\textwidth}{|X|X|X|X}
        Stable Diffusion & Ours & Stable Diffusion & Ours
    \end{tabularx}
    \begin{subfigure}[b]{0.495\textwidth}
        \centering
        \includegraphics[width=0.495\textwidth]{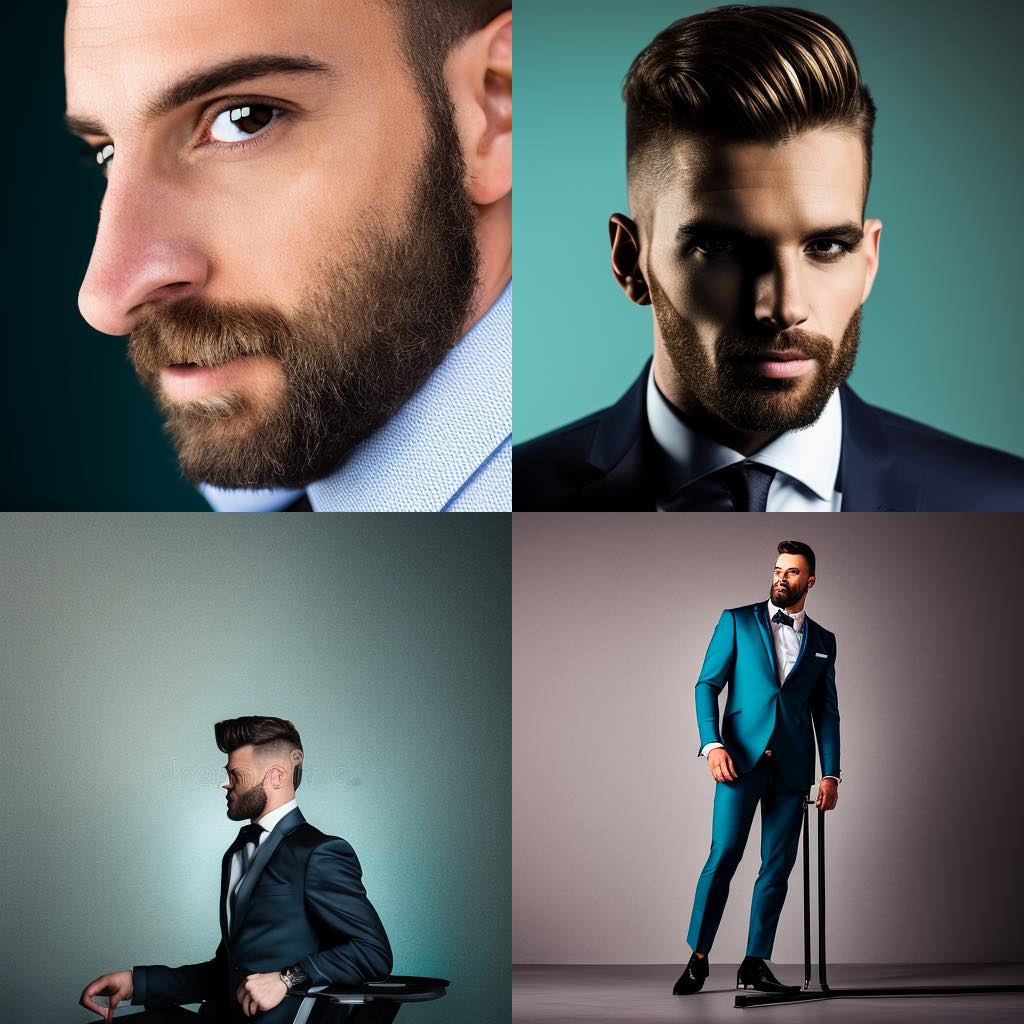}
        \includegraphics[width=0.495\textwidth]{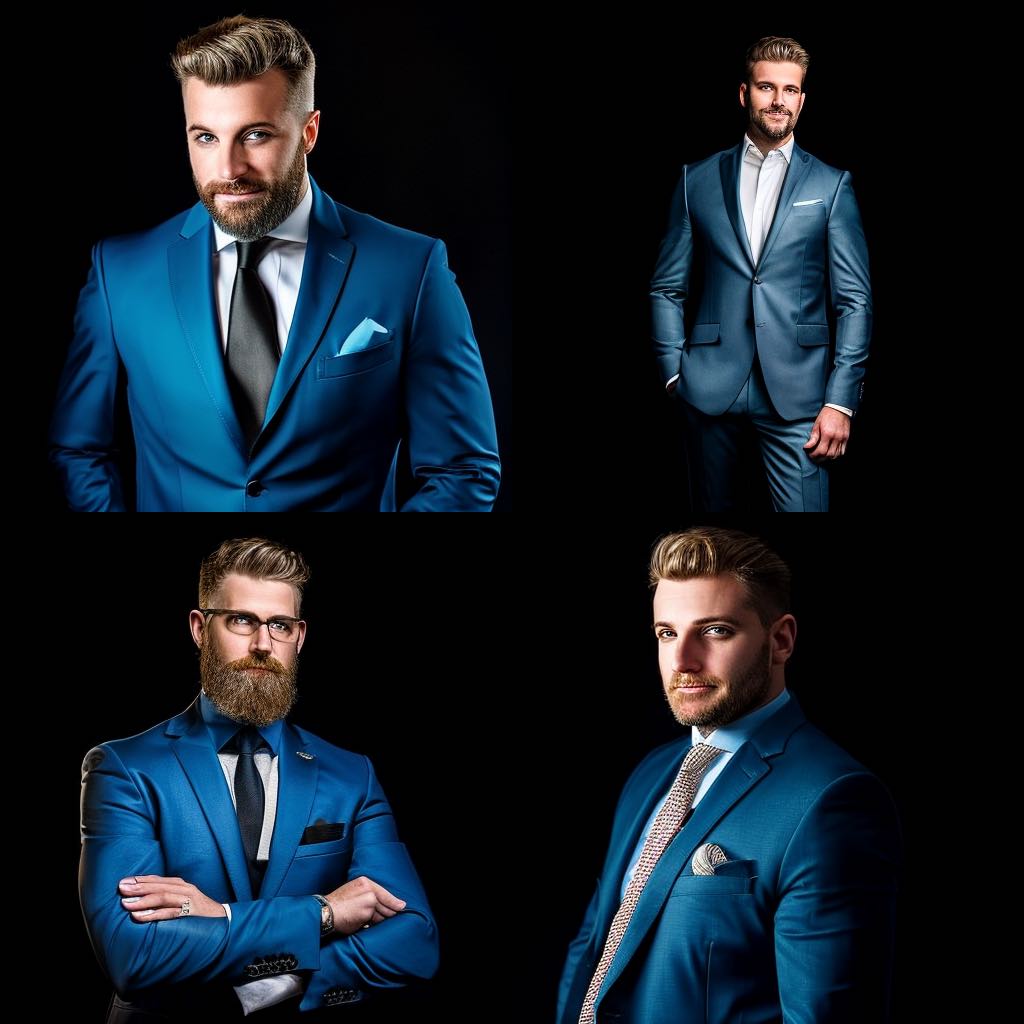}
        \caption{Close-up portrait of a man wearing suit posing in a dark studio, rim lighting, teal hue, octane, unreal}
    \end{subfigure}
    \begin{subfigure}[b]{0.495\textwidth}
        \centering
        \includegraphics[width=0.495\textwidth]{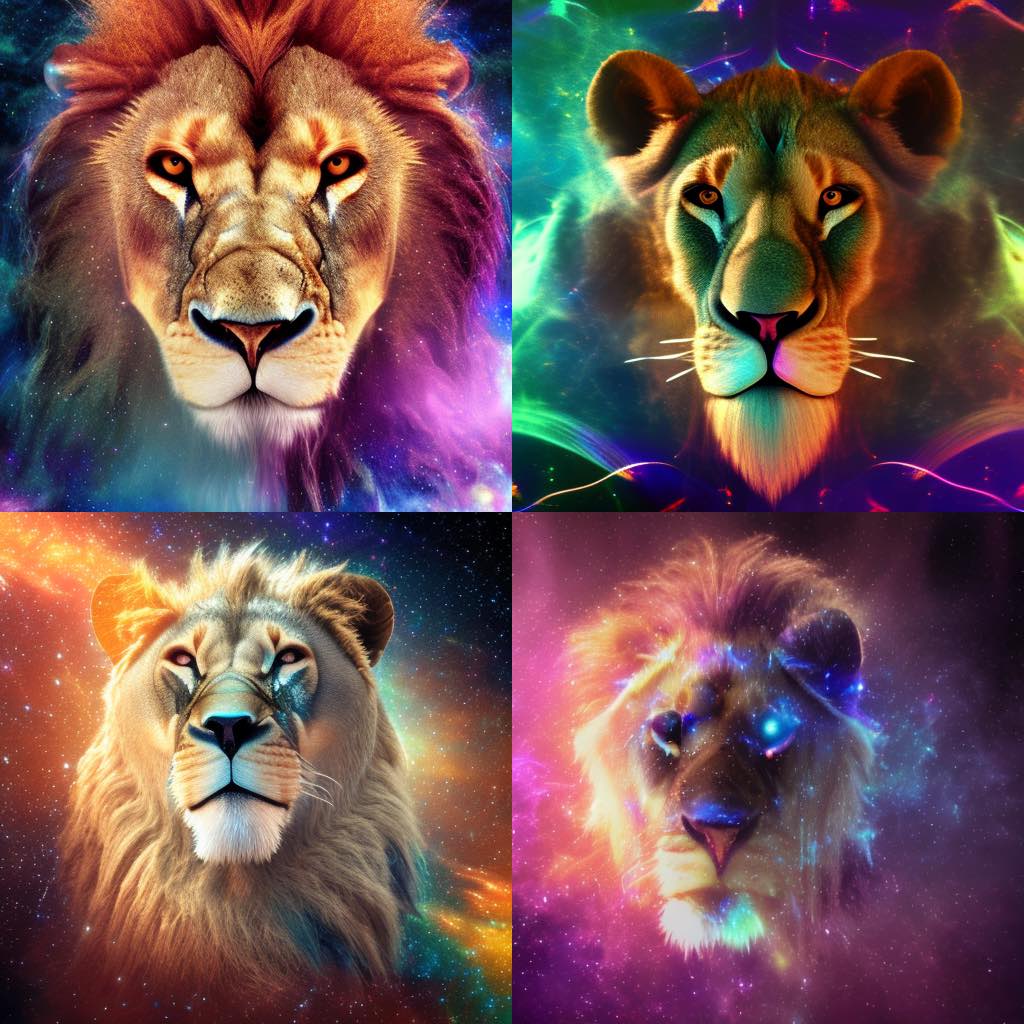}
        \includegraphics[width=0.495\textwidth]{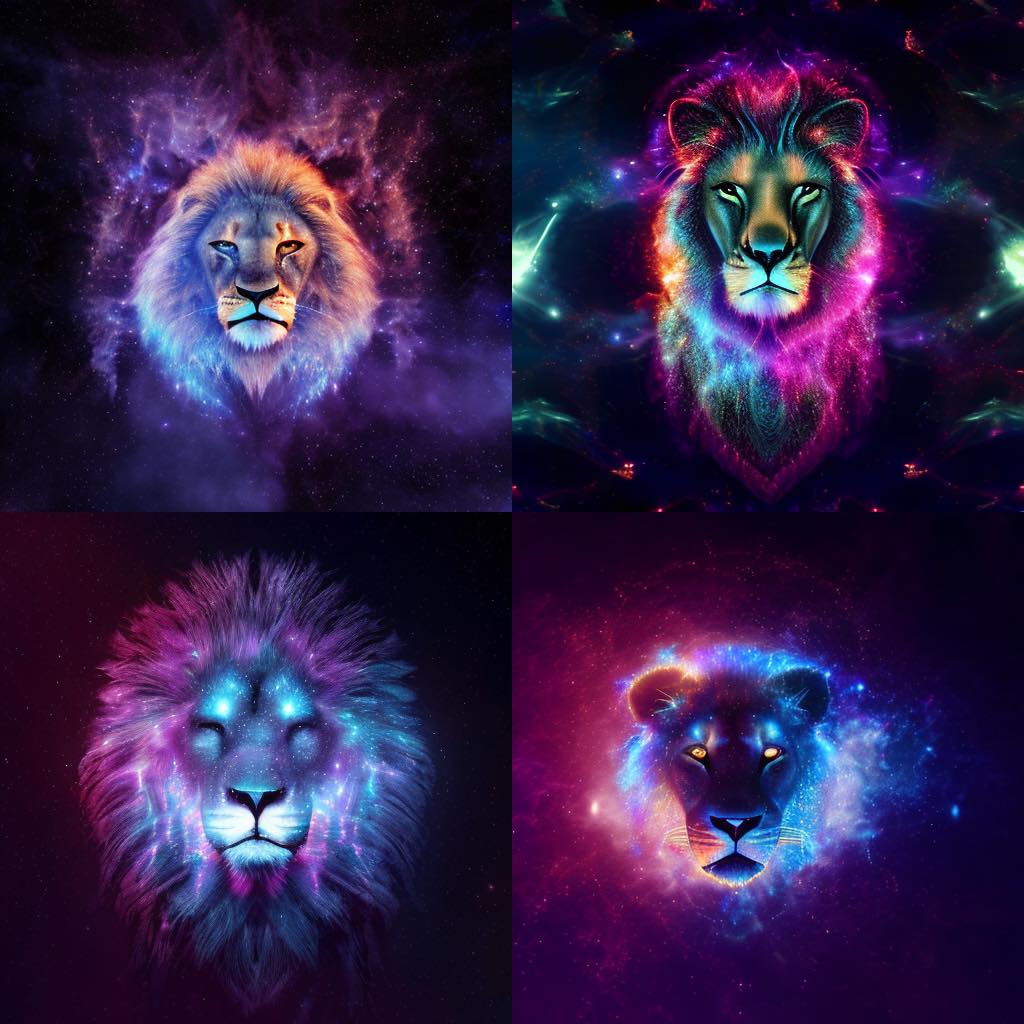}
        \caption{A lion in galaxies, spirals, nebulae, stars, smoke, iridescent, intricate detail, octane render, 8k}
    \end{subfigure}
    \vspace{1em}
    \begin{subfigure}[b]{0.495\textwidth}
        \centering
        \includegraphics[width=0.495\textwidth]{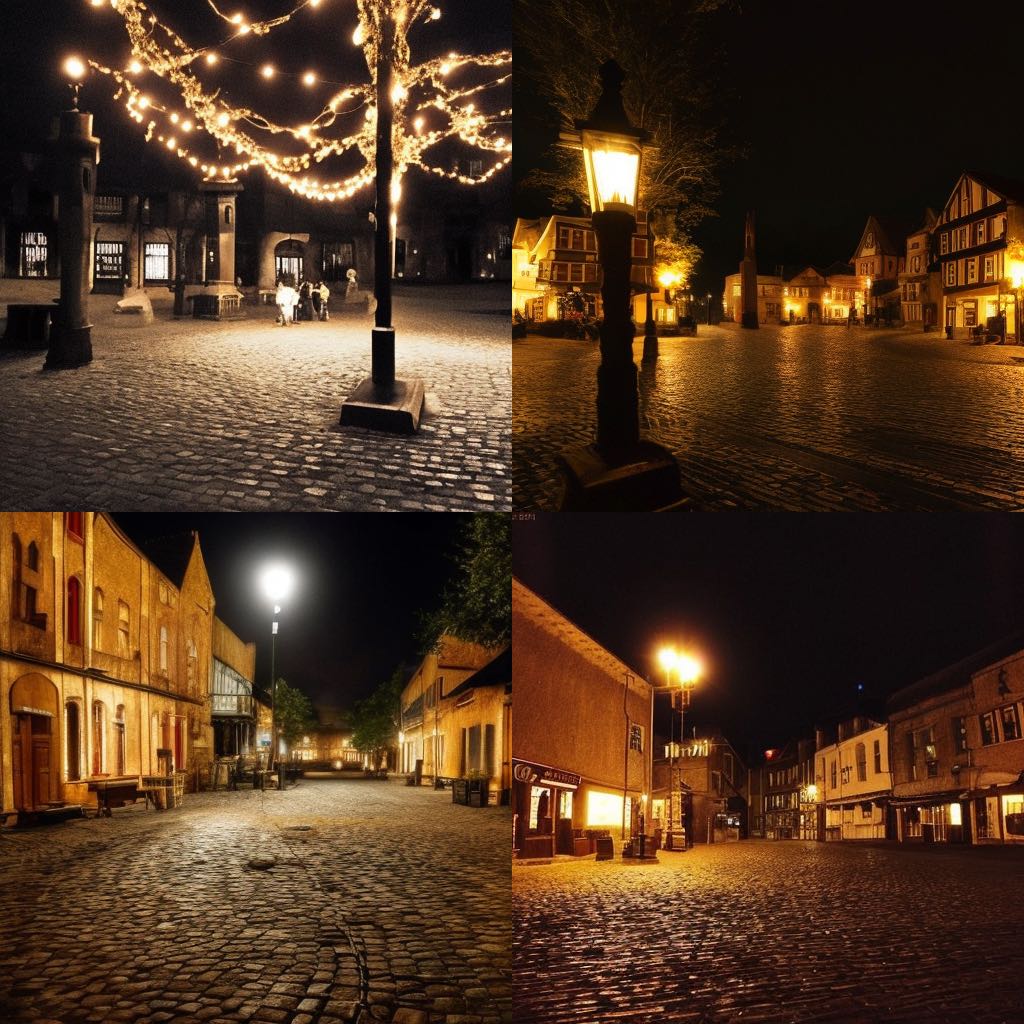}
        \includegraphics[width=0.495\textwidth]{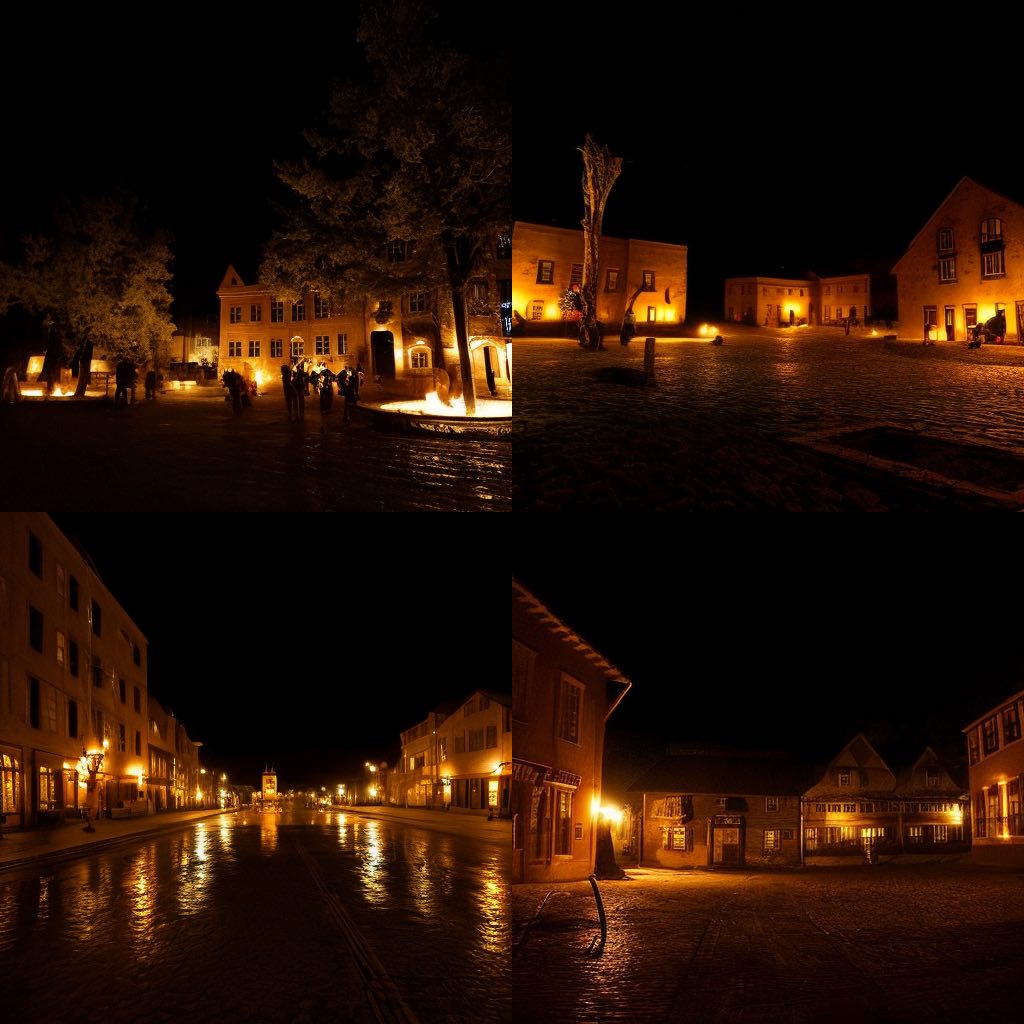}
        \caption{A dark town square lit only by a few torchlights}
    \end{subfigure}
    \begin{subfigure}[b]{0.495\textwidth}
        \centering
        \includegraphics[width=0.495\textwidth]{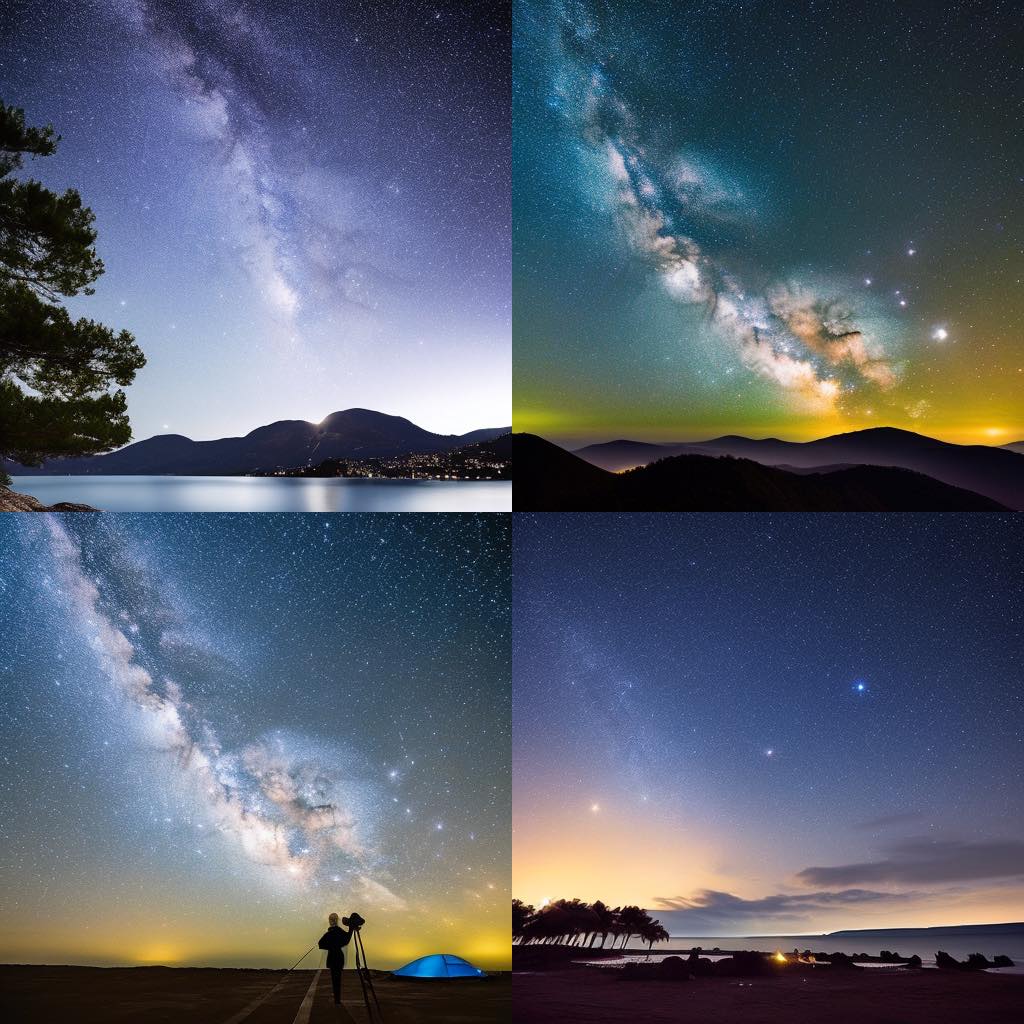}
        \includegraphics[width=0.495\textwidth]{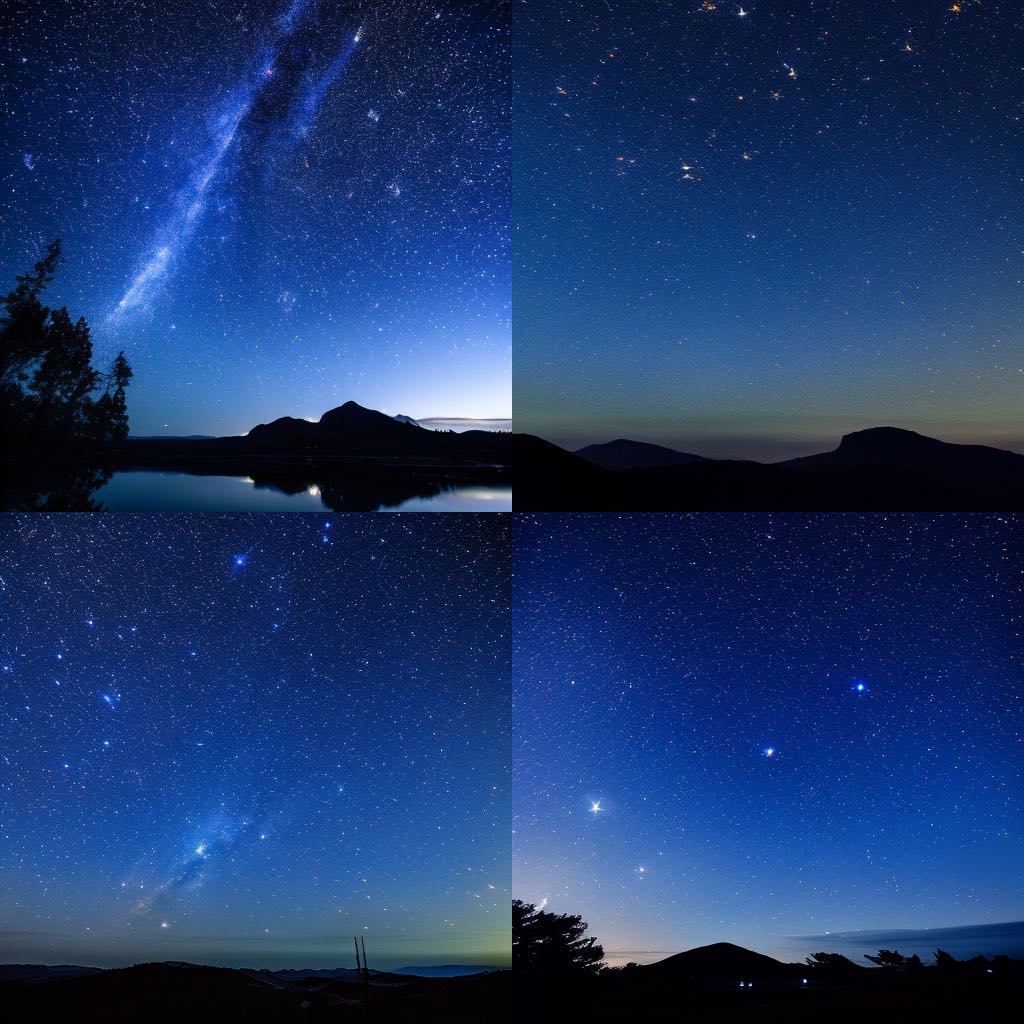}
        \caption{A starry sky}
    \end{subfigure}
    \vspace{1em}
    \begin{subfigure}[b]{0.495\textwidth}
        \centering
        \includegraphics[width=0.495\textwidth]{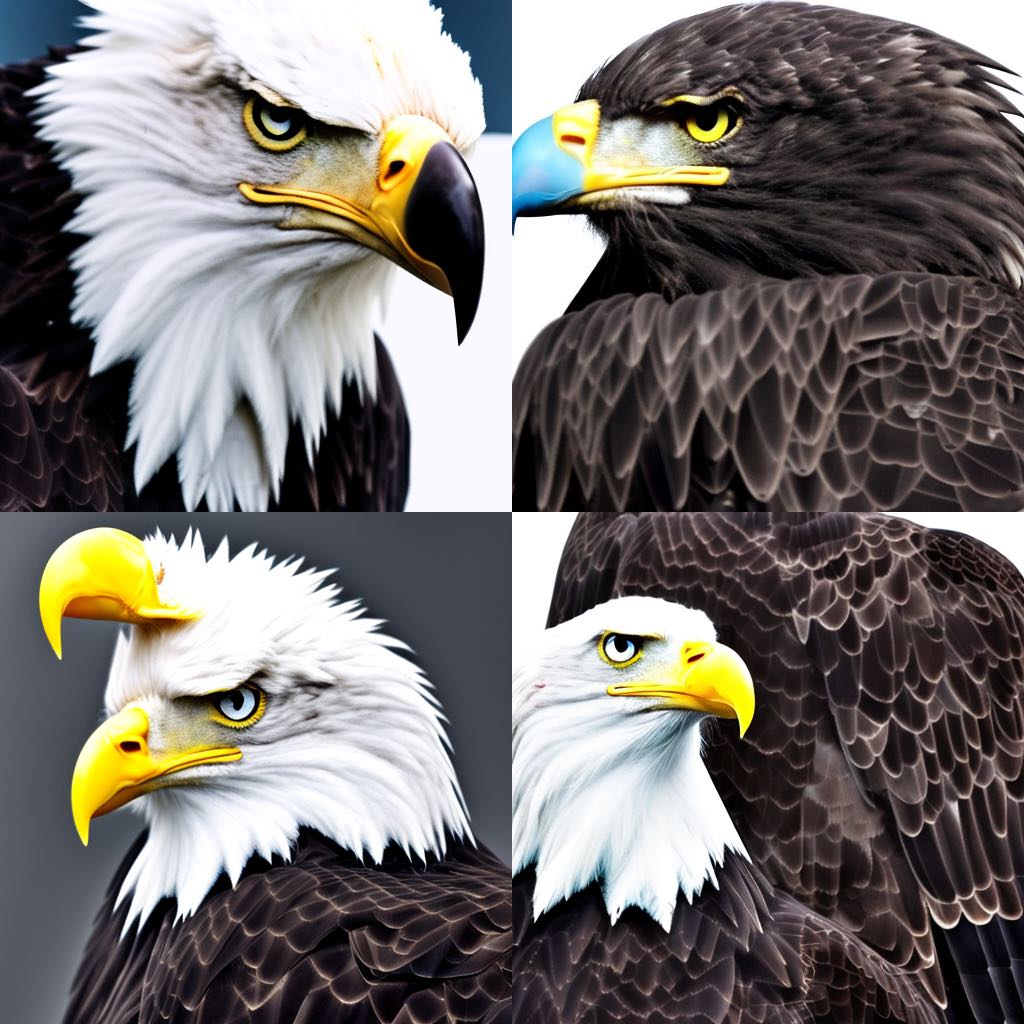}
        \includegraphics[width=0.495\textwidth]{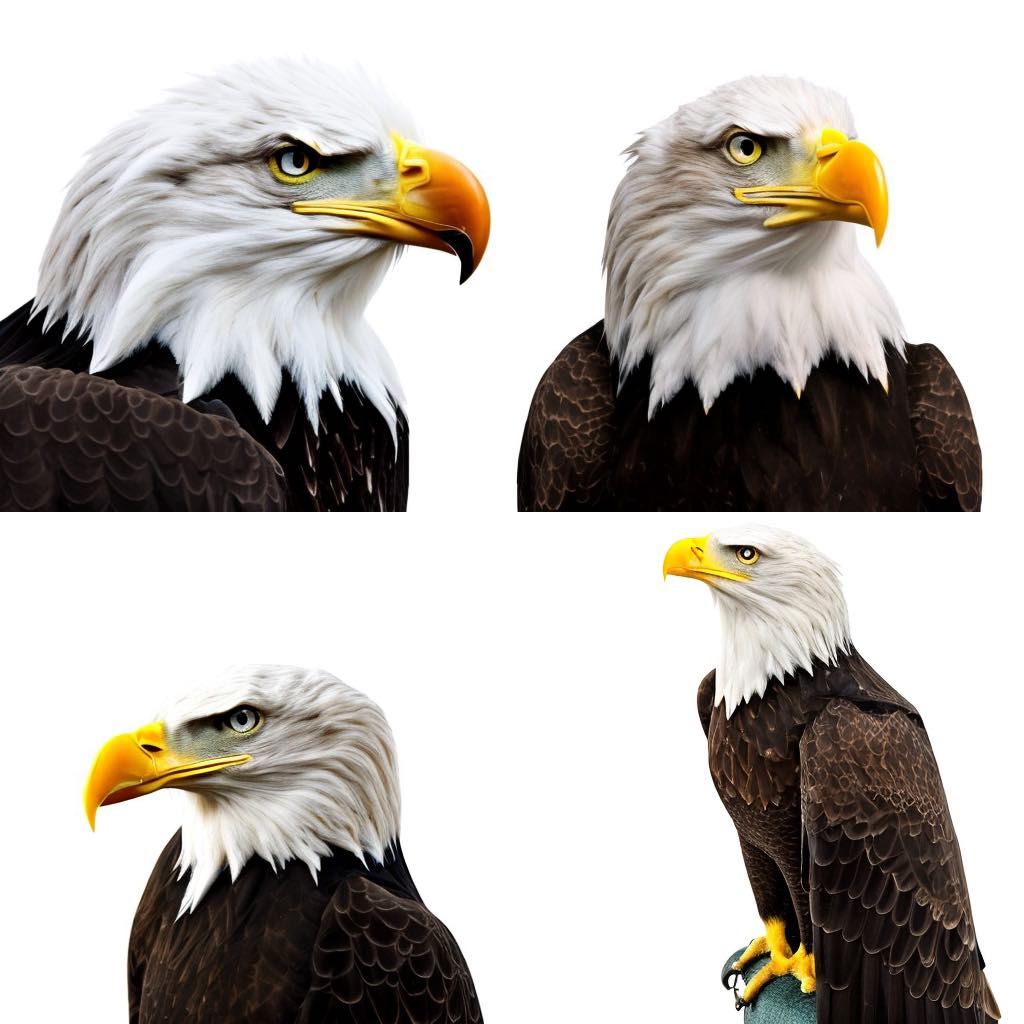}
        \caption{A bald eagle against a white background}
    \end{subfigure}
    \begin{subfigure}[b]{0.495\textwidth}
        \centering
        \includegraphics[width=0.495\textwidth]{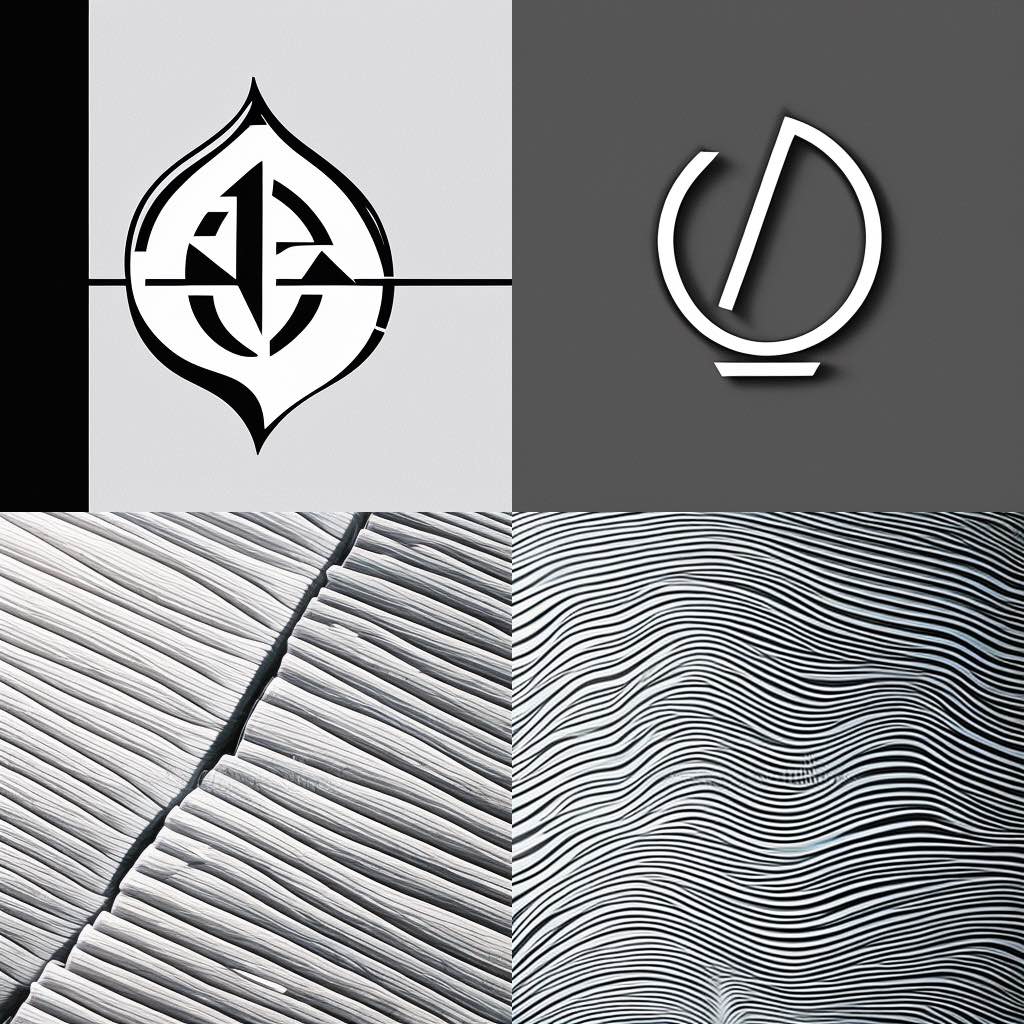}
        \includegraphics[width=0.495\textwidth]{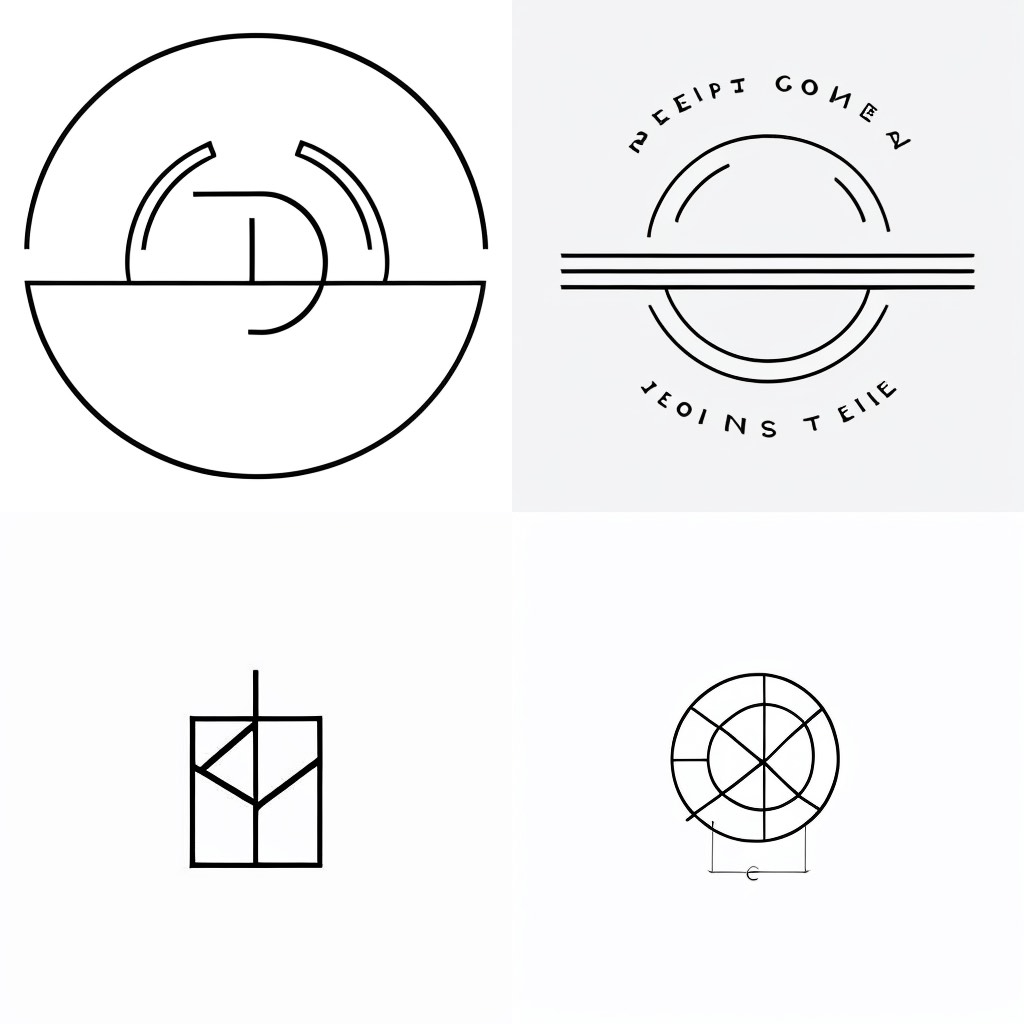}
        \caption{Monochrome line-art logo on a white background}
    \end{subfigure}
    \begin{subfigure}[b]{0.495\textwidth}
        \centering
        \includegraphics[width=0.495\textwidth]{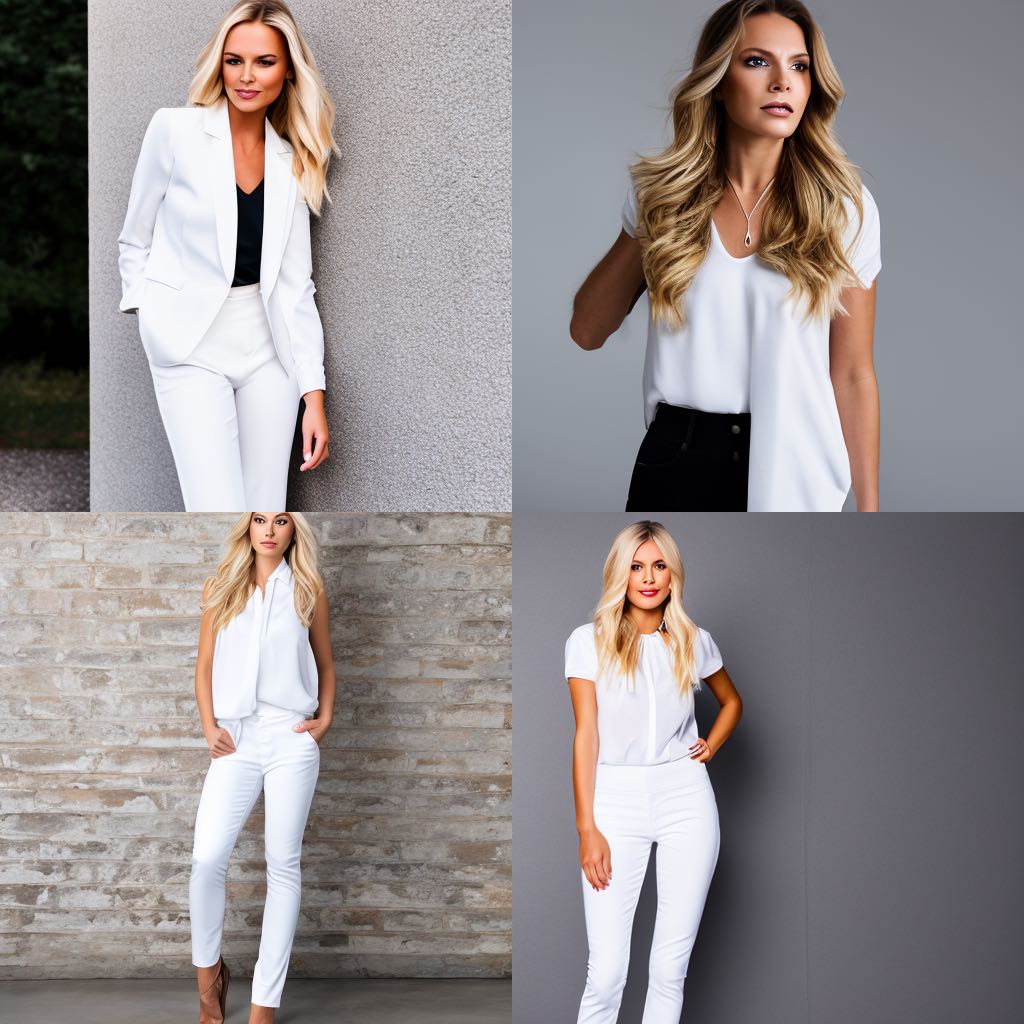}
        \includegraphics[width=0.495\textwidth]{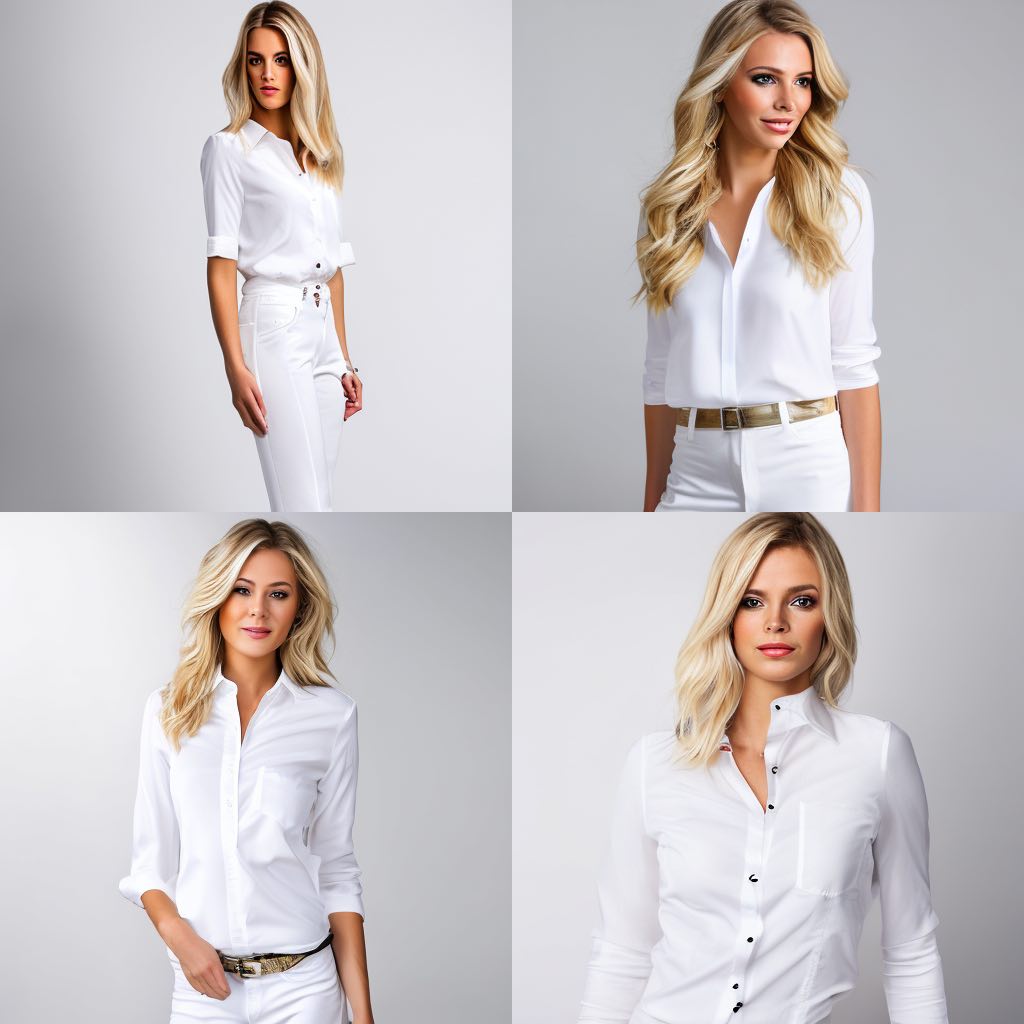}
        \caption{Blonde female model, white shirt, white pants, white background, studio}
    \end{subfigure}
    \begin{subfigure}[b]{0.495\textwidth}
        \centering
        \includegraphics[width=0.495\textwidth]{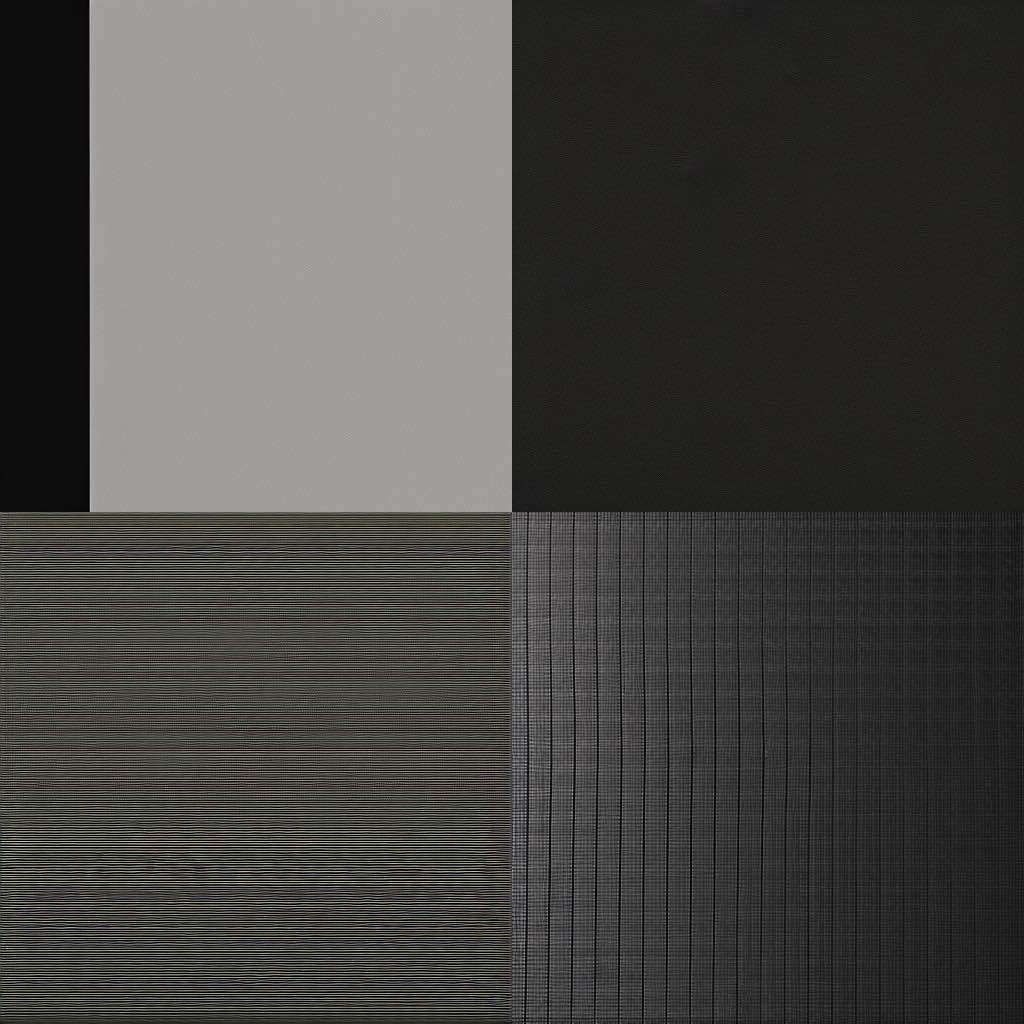}
        \includegraphics[width=0.495\textwidth]{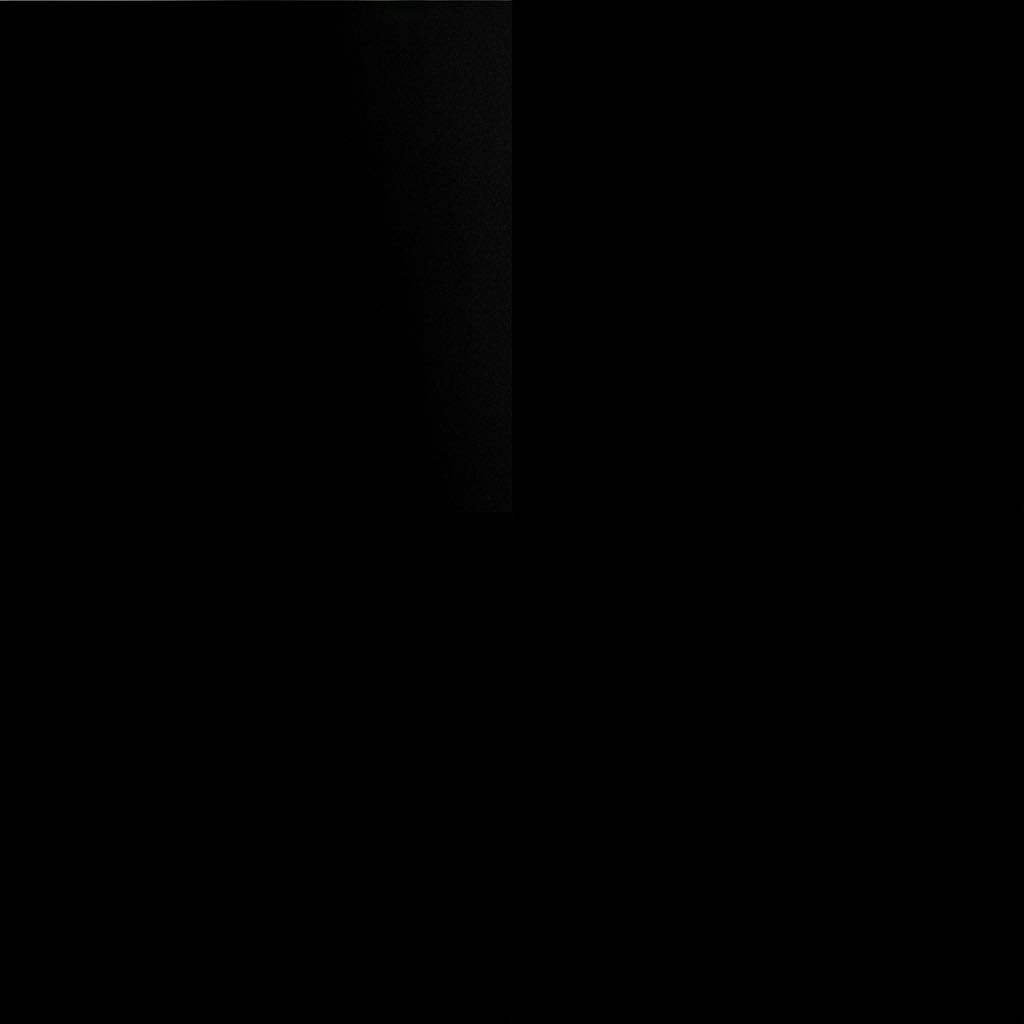}
        \caption{Solid black background}
    \end{subfigure}
    \caption{Qualitative comparison. Left is Stable Diffusion reference model. Right is Stable Diffusion model after applying all the proposed fixes. All images are produced using DDIM sampler, $S=50$ steps, trailing timestep selection, classifier-free guidance weight $w=7.5$, and rescale factor $\phi=0.7$. Images within a pair are generated using the same seed. Different negative prompts are used.}
    \label{fig:qualitative}
\end{figure*}

\Cref{fig:qualitative} shows our method can generate images with a diverse brightness range. Specifically, the model with flawed designs always generates samples with medium brightness. It is unable to generate correct images when given explicit prompts, such as ``white background'' and ``Solid black background'', etc. In contrast, our model is able to generate according to the prompts perfectly.

\subsection{Quantitative}

We follow the convention to calculate Fréchet Inception Distance (FID) \cite{heusel2018gans,parmar2022aliased} and Inception Score (IS) \cite{salimans2016improved}. We randomly select 10k images from COCO 2014 validation dataset \cite{lin2015microsoft} and use our models to generate with the corresponding captions. \Cref{tab:fid} shows that our model has improved FID/IS, suggesting our model better fits the image distribution and are more visually appealing.

\begin{table}[h]
    \centering
    \begin{tabularx}{\linewidth}{Xrr}
        \toprule
        Model & FID $\downarrow$ & IS $\uparrow$ \\
        \midrule
        SD v2.1-base official & 23.76 & 32.84 \\
        SD with our data, no fixes & 22.96 & 34.11 \\
        \textbf{SD with fixes (Ours)} & \textbf{21.66} & \textbf{36.16} \\
        \bottomrule
    \end{tabularx}
    \vspace{-0.2em}
    \caption{Quantitative evaluation. All models use DDIM sampler with $S=50$ steps, guidance weight $w=7.5$, and no negative prompt. Ours uses zero terminal SNR noise schedule, v prediction, trailing sample steps, and guidance rescale factor $\phi=0.7$.}
    \label{tab:fid}
\end{table}

\section{Ablation}

\subsection{Comparison of Sample Steps}

\begin{figure}[h]
    \centering
    \begin{subfigure}[b]{0.15\textwidth}
         \centering
         \includegraphics[width=\textwidth]{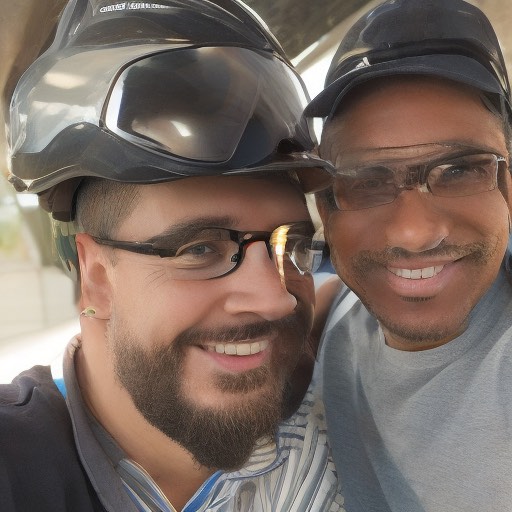}
         \caption{Leading, $S=5$}
    \end{subfigure}
    \begin{subfigure}[b]{0.15\textwidth}
         \centering
         \includegraphics[width=\textwidth]{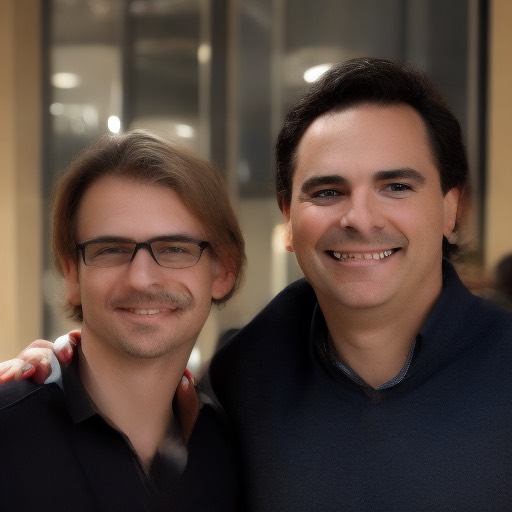}
         \caption{Linspace, $S=5$}
    \end{subfigure}
    \begin{subfigure}[b]{0.15\textwidth}
         \centering
         \includegraphics[width=\textwidth]{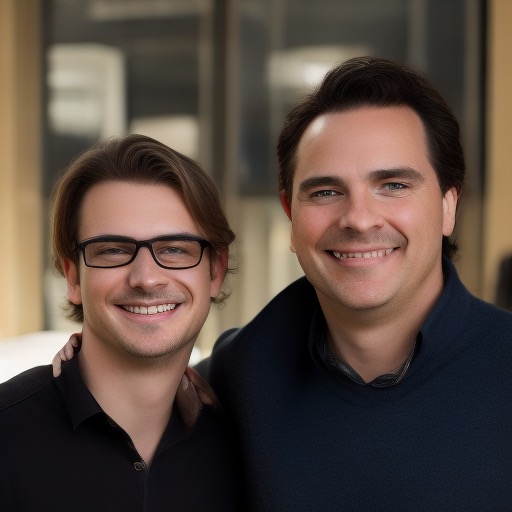}
         \caption{Trailing, $S=5$}
    \end{subfigure}
    \begin{subfigure}[b]{0.15\textwidth}
         \centering
         \includegraphics[width=\textwidth]{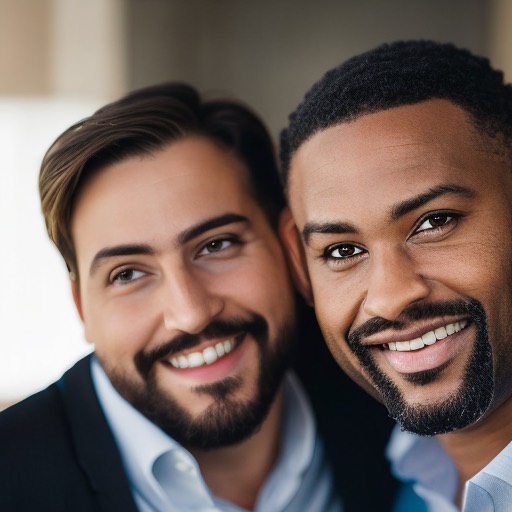}
         \caption{Leading, $S=25$}
    \end{subfigure}
    \begin{subfigure}[b]{0.15\textwidth}
         \centering
         \includegraphics[width=\textwidth]{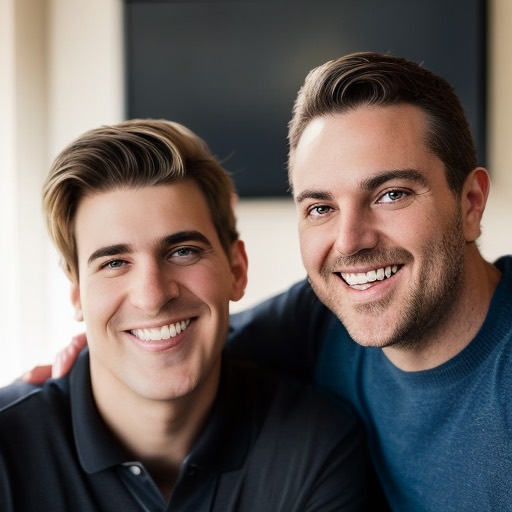}
         \caption{Linspace, $S=25$}
    \end{subfigure}
    \begin{subfigure}[b]{0.15\textwidth}
         \centering
         \includegraphics[width=\textwidth]{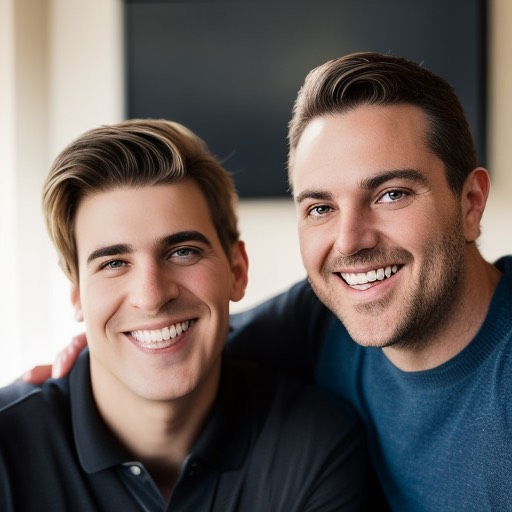}
         \caption{Trailing, $S=25$}
    \end{subfigure}
    \caption{Comparison of sample steps selections. The prompt is: ``A close-up photograph of two men smiling in bright light''. Sampled with DDIM. Same seed. When the sample step is extremely small, \eg $S=5$, trailing is noticeably better than linspace. When the sample step is large, \eg $S=25$, the difference between trailing and linspace is subtle.}
    \label{fig:ablation_sample_steps}
\end{figure}

\Cref{fig:ablation_sample_steps} compares sampling using leading, linspace, and trailing on our model trained with zero terminal SNR noise schedule. When sample step $S$ is small, \eg taking $S=5$ as an extreme example, Trailing noticeably outperforms linspace. But for common choices such as $S=25$, the difference between trailing and linspace is not easily noticeable.

\subsection{Analyzing Model Behavior with Zero SNR}

Let's consider an ``ideal" unconditional model that has been trained till perfect convergence with zero terminal SNR. At $t=T$, the model learns to predict the same exact L2 mean of all the data samples regardless of the noise input. In the text-conditional case, the model will predict the L2 mean conditioned on the prompt but invariant to the noise input.

Therefore, the first sample step at $t=T$ ideally yields the same exact prediction regardless of noise input. The variation begins on the second sample step. In DDPM \cite{ho2020denoising}, different random Gaussian noise is added back to the same predicted $x_0$ from the first step. In DDIM \cite{song2022denoising}, different predicted noise is added back to the same predicted $x_0$ from the first step. The posterior probability for $x_0$ now becomes different and the model starts to generate different results on different noised inputs.

\begin{figure*}[h]
    \includegraphics[width=\linewidth]{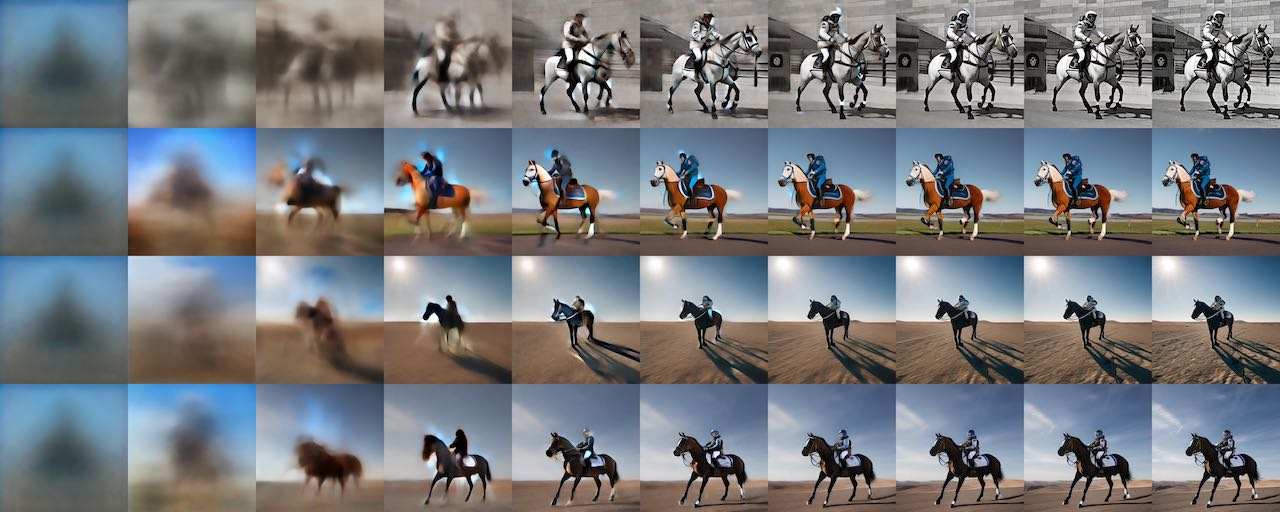}
    \setlength{\tabcolsep}{0em}
    \footnotesize
    \newcolumntype{Y}{>{\centering\arraybackslash}X}
    \begin{tabularx}{\linewidth}{YYYYYYYYYY}
        1000 & 900 & 800 & 700 & 600 & 500 & 400 & 300 & 200 & 100
    \end{tabularx}
    \caption{Visualization of the sample steps on prompt ``An astronaut riding a horse''. Horizontal axis is the timestep $t$. At $t=T$, the model generates the mean of the data distribution based on the prompt.}
    \label{fig:ablation_steps_visualize}
\end{figure*}

This is congruent with our model behavior. \Cref{fig:ablation_steps_visualize} shows that our model predicts almost exact results regardless of the noise input at $t=T$, and the variation begins from the next sample step.

In another word, at $t=T$, the noise input is unnecessary, except we keep it for architectural convenience.

\subsection{Effect of Classifier-Free Guidance Rescale}

\begin{figure*}[h]

    \setlength{\tabcolsep}{0.1em}
    \footnotesize
    \begin{tabularx}{\textwidth}{ccccc}

        \includegraphics[width=0.196\textwidth]{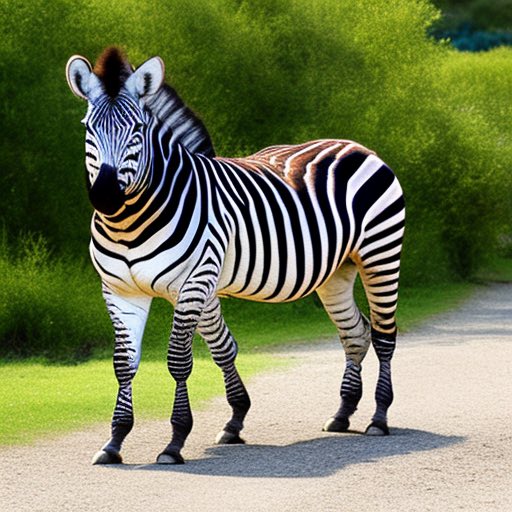} &
        \includegraphics[width=0.196\textwidth]{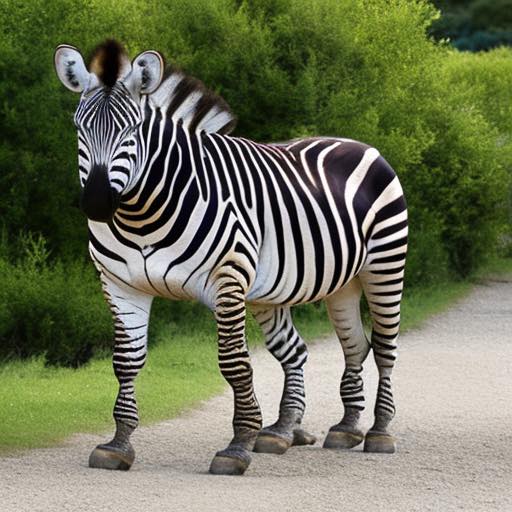} &
        \includegraphics[width=0.196\textwidth]{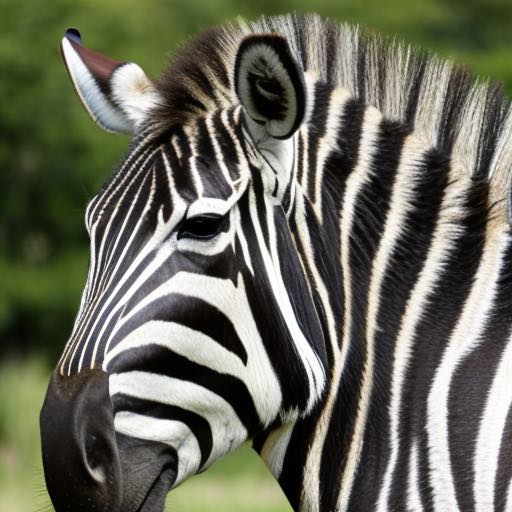} &
        \includegraphics[width=0.196\textwidth]{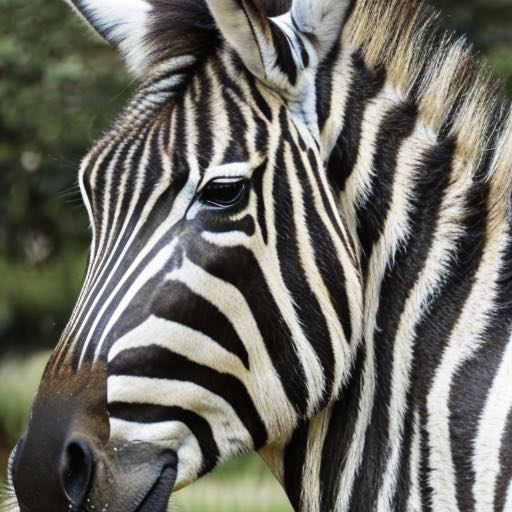} &
        \includegraphics[width=0.196\textwidth]{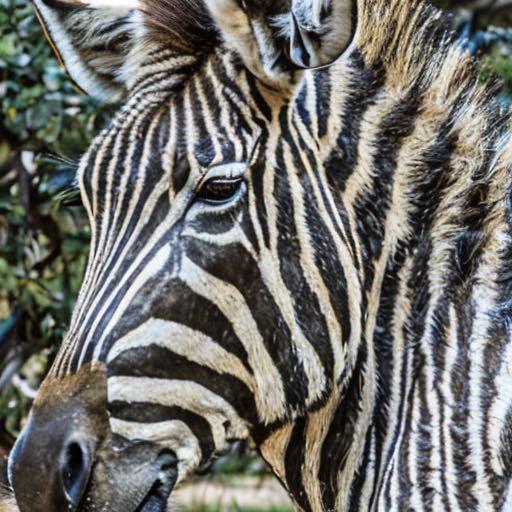} \\

        \includegraphics[width=0.196\textwidth]{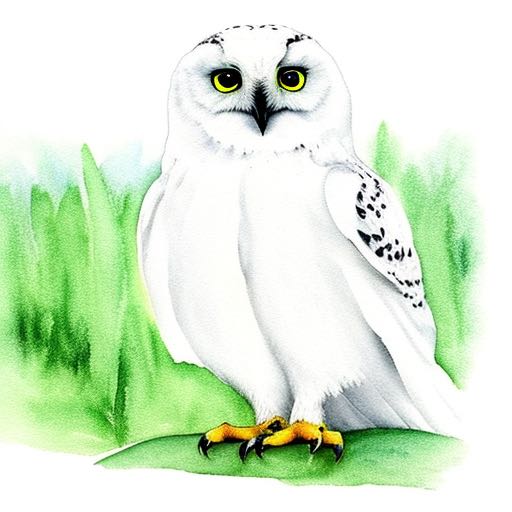} &
        \includegraphics[width=0.196\textwidth]{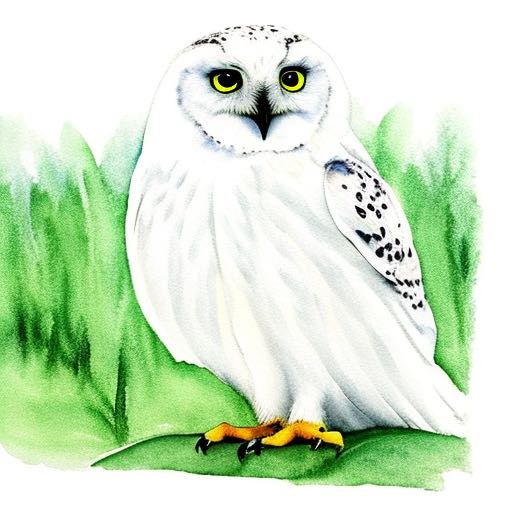} &
        \includegraphics[width=0.196\textwidth]{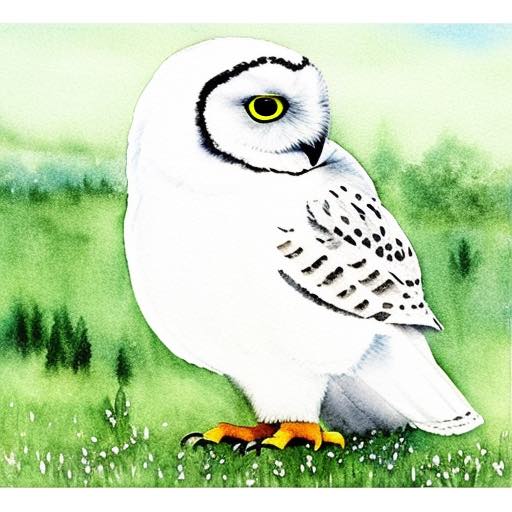} &
        \includegraphics[width=0.196\textwidth]{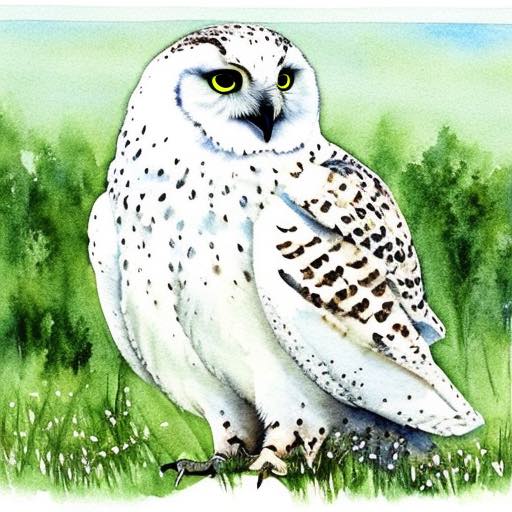} &
        \includegraphics[width=0.196\textwidth]{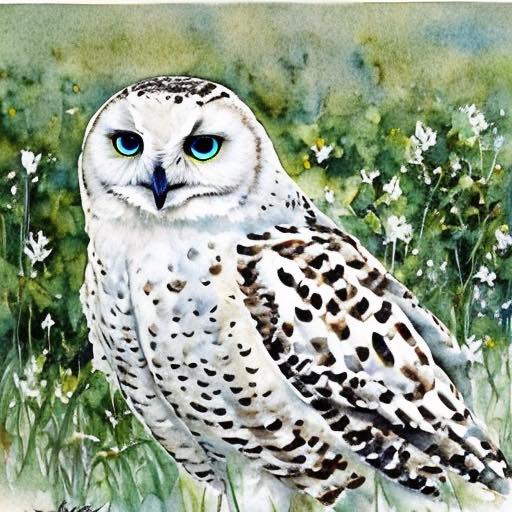} \\

        \includegraphics[width=0.196\textwidth]{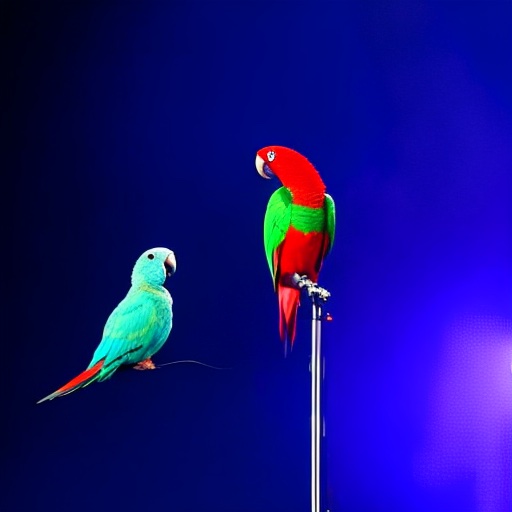} &
        \includegraphics[width=0.196\textwidth]{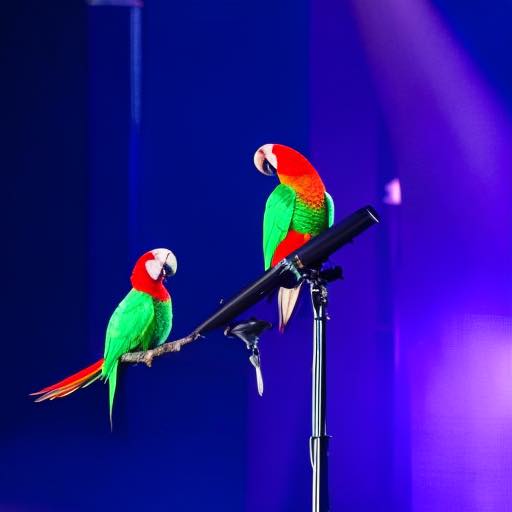} &
        \includegraphics[width=0.196\textwidth]{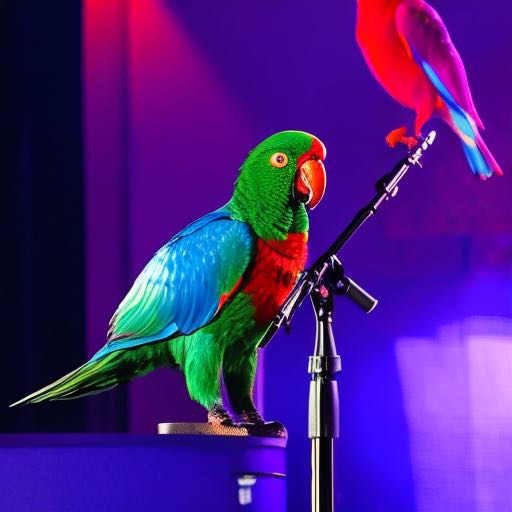} &
        \includegraphics[width=0.196\textwidth]{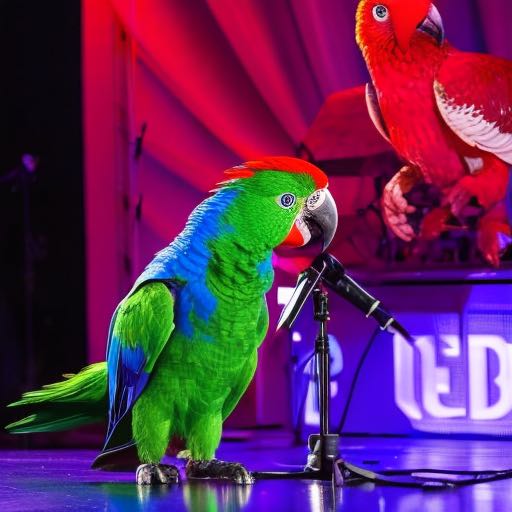} &
        \includegraphics[width=0.196\textwidth]{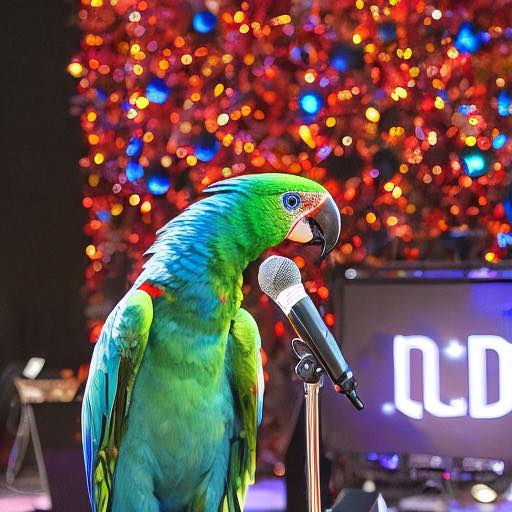} \\
        
        $\phi=0$ & $\phi=0.25$ & $\phi=0.5$ & $\phi=0.75$ & $\phi=1$
    \end{tabularx}
    \vspace{-5px}
    \caption{Comparison of classifier-free guidance rescale factor $\phi$. All images use DDIM sampler with $S=25$ steps and guidance weight $w=7.5$. Regular classifier-free guidance is equivalent to $\phi = 0$ and can cause over-exposure. We find $\phi \in [0.5,\dots,0.75]$ to work well. The positive prompts are (1) ``A zebra'', (2) ``A watercolor painting of a snowy owl standing in a grassy field'', (3) ``A photo of a red parrot, a blue parrot and a green parrot singing at a concert in front of a microphone. Colorful lights in the background.''. Different negative prompts are used.}
    \label{fig:ablation_cfg_rescale}
\end{figure*}

\Cref{fig:ablation_cfg_rescale} compares the results of using different rescale factors $\phi$. When using regular classifier-free guidance, corresponding to rescale factor $\phi=0$, the images tend to overexpose. We empirically find that setting $\phi$ to be within 0.5 and 0.75 produces the most appealing results.

\subsection{Comparison to Offset Noise}

Offset noise is another technique proposed in \cite{offsetnoise} to address the brightness problem in Stable Diffusion. Instead of sampling $\epsilon \sim \mathcal{N}(0, \textbf{I})$, they propose to sample $\epsilon_{hwc} \sim \mathcal{N}(0.1\delta_c, \textbf{I})$, where $\delta_c \sim \mathcal{N}(0, \textbf{I})$ and the same $\delta_c$ is used for every pixel $h,w$ in every channel $c$.

When using offset noise, the noise at each pixel is no longer iid. since $\delta_c$ shifts the entire channel together. The mean of the noised input is no longer indicative of the mean of the true image. Therefore, the model learns not to respect the mean of its input when predicting the output at all timesteps. So even if pure Gaussian noise is given at $t=T$ and the signal is leaked by the flawed noise schedule, the model ignores it and is free to alter the output mean at every timestep.

Offset noise does enable Stable Diffusion model to generate very bright and dark samples but it is incongruent with the theory of the diffusion process and may generate samples with brightness that does not fit the true data distribution, i.e. too bright or too dark. It is a trick that does not address the fundamental issue.

\section{Implementation}
\label{sec:implementation}

In this section, we show zero terminal SNR is valid from diffusion's math perspective and point out common pitfalls in sampler implementations.

Implementations of samplers must avoid $\epsilon$ math formulation. Consider DDPM \cite{ho2020denoising} sampling. Some implementations handle $v$ prediction by first converting it to $\epsilon$ using \Cref{eq:v_to_eps}, then applying sampling equation \Cref{eq:ddpm_eps_formulation} (Equation 11 in \cite{ho2020denoising}). This is problematic for zero SNR at the terminal step, because the conversion to $\epsilon$ loses all the signal information and $\alpha_t=0$ causes zero division error.

\begin{equation}
\epsilon = \sqrt{\bar\alpha_t} v + \sqrt{1 - \bar\alpha_t} x_t
\label{eq:v_to_eps}
\end{equation}

\begin{equation}
\Tilde{\mu}_t := \frac{1}{\sqrt{\alpha_t}} (x_t - \frac{\beta_t}{\sqrt{1 - \bar{\alpha}_t}} \epsilon)
\label{eq:ddpm_eps_formulation}
\end{equation}

The correct way is to first convert $v$ prediction to $x_0$ in \Cref{eq:v_to_x0}, then sample directly with $x_0$ formulation as in \Cref{eq:ddpm_x0_formulation} (Equation 7 in \cite{ho2020denoising}). This avoids the singularity problem in $\epsilon$ formulation.

\begin{equation}
x_0 = \sqrt{\bar\alpha_t} x_t - \sqrt{1 - \bar\alpha_t} v
\label{eq:v_to_x0}
\end{equation}

\begin{equation}
\Tilde{\mu}_t := \frac{\sqrt{\bar{\alpha}_{t-1}}\beta_t}{1 - \bar{\alpha}_t} x_0 + \frac{\sqrt{\alpha_t}(1-\bar{\alpha}_{t-1})}{1-\bar{\alpha}_t} x_t
\label{eq:ddpm_x0_formulation}
\end{equation}

For DDIM \cite{song2022denoising}, first convert $v$ prediction to $\epsilon, x_0$ with \Cref{eq:v_to_eps,eq:v_to_x0} then sample with \Cref{eq:ddim} (Equation 12 in \cite{song2022denoising}):
\begin{equation}
x_{t-1} = \sqrt{\bar\alpha_{t-1}} x_0 + \sqrt{1 - \bar\alpha_{t-1} - \sigma^2_t} \epsilon + \sigma_t z
\label{eq:ddim}
\end{equation}
where $z \sim \mathcal{N}(0, \mathbf{I})$, $\eta \in [0, 1]$, and
\begin{equation}
\sigma_t(\eta) = \eta \sqrt{\frac{1 - \bar\alpha_{t-1}}{1-\bar\alpha_t}} \sqrt{1-\frac{\bar\alpha_t}{\bar\alpha_{t-1}}}
\end{equation}

Same logic applies to other samplers.

\section{Conclusion}

In summary, we have discovered that diffusion models should use noise schedules with zero terminal SNR and should be sampled starting from the last timestep in order to ensure the training behavior is aligned with inference. We have proposed a simple way to rescale existing noise schedules to enforce zero terminal SNR and a classifier-free guidance rescaling technique to counter image over-exposure. We encourage future designs of diffusion models to take this into account.

\newpage

%%%%%%%%% REFERENCES
{\small
\bibliographystyle{ieee_fullname}
\bibliography{egbib}

\begin{thebibliography}{10}\itemsep=-1pt

\bibitem{offsetnoise}
Nicholas Guttenberg.
\newblock Diffusion with offset noise, 2023.

\bibitem{heusel2018gans}
Martin Heusel, Hubert Ramsauer, Thomas Unterthiner, Bernhard Nessler, and Sepp
  Hochreiter.
\newblock Gans trained by a two time-scale update rule converge to a local nash
  equilibrium, 2018.

\bibitem{ho2020denoising}
Jonathan Ho, Ajay Jain, and Pieter Abbeel.
\newblock Denoising diffusion probabilistic models, 2020.

\bibitem{ho2022classifierfree}
Jonathan Ho and Tim Salimans.
\newblock Classifier-free diffusion guidance, 2022.

\bibitem{lin2015microsoft}
Tsung-Yi Lin, Michael Maire, Serge Belongie, Lubomir Bourdev, Ross Girshick,
  James Hays, Pietro Perona, Deva Ramanan, C.~Lawrence Zitnick, and Piotr
  Dollár.
\newblock Microsoft coco: Common objects in context, 2015.

\bibitem{liu2022pseudo}
Luping Liu, Yi Ren, Zhijie Lin, and Zhou Zhao.
\newblock Pseudo numerical methods for diffusion models on manifolds.
\newblock In {\em International Conference on Learning Representations}, 2022.

\bibitem{lu2022dpmsolver}
Cheng Lu, Yuhao Zhou, Fan Bao, Jianfei Chen, Chongxuan Li, and Jun Zhu.
\newblock Dpm-solver: A fast ode solver for diffusion probabilistic model
  sampling in around 10 steps, 2022.

\bibitem{nichol2021improved}
Alex Nichol and Prafulla Dhariwal.
\newblock Improved denoising diffusion probabilistic models, 2021.

\bibitem{parmar2022aliased}
Gaurav Parmar, Richard Zhang, and Jun-Yan Zhu.
\newblock On aliased resizing and surprising subtleties in gan evaluation,
  2022.

\bibitem{rombach2021highresolution}
Robin Rombach, Andreas Blattmann, Dominik Lorenz, Patrick Esser, and Björn
  Ommer.
\newblock High-resolution image synthesis with latent diffusion models, 2021.

\bibitem{saharia2022photorealistic}
Chitwan Saharia, William Chan, Saurabh Saxena, Lala Li, Jay Whang, Emily
  Denton, Seyed Kamyar~Seyed Ghasemipour, Burcu~Karagol Ayan, S.~Sara Mahdavi,
  Rapha~Gontijo Lopes, Tim Salimans, Jonathan Ho, David~J Fleet, and Mohammad
  Norouzi.
\newblock Photorealistic text-to-image diffusion models with deep language
  understanding, 2022.

\bibitem{salimans2016improved}
Tim Salimans, Ian Goodfellow, Wojciech Zaremba, Vicki Cheung, Alec Radford, and
  Xi Chen.
\newblock Improved techniques for training gans, 2016.

\bibitem{salimans2022progressive}
Tim Salimans and Jonathan Ho.
\newblock Progressive distillation for fast sampling of diffusion models, 2022.

\bibitem{schuhmann2022laion5b}
Christoph Schuhmann, Romain Beaumont, Richard Vencu, Cade Gordon, Ross
  Wightman, Mehdi Cherti, Theo Coombes, Aarush Katta, Clayton Mullis, Mitchell
  Wortsman, Patrick Schramowski, Srivatsa Kundurthy, Katherine Crowson, Ludwig
  Schmidt, Robert Kaczmarczyk, and Jenia Jitsev.
\newblock Laion-5b: An open large-scale dataset for training next generation
  image-text models, 2022.

\bibitem{sohldickstein2015deep}
Jascha Sohl-Dickstein, Eric~A. Weiss, Niru Maheswaranathan, and Surya Ganguli.
\newblock Deep unsupervised learning using nonequilibrium thermodynamics, 2015.

\bibitem{song2022denoising}
Jiaming Song, Chenlin Meng, and Stefano Ermon.
\newblock Denoising diffusion implicit models, 2022.

\bibitem{song2021scorebased}
Yang Song, Jascha Sohl-Dickstein, Diederik~P. Kingma, Abhishek Kumar, Stefano
  Ermon, and Ben Poole.
\newblock Score-based generative modeling through stochastic differential
  equations, 2021.

\end{thebibliography}
}

\end{document}